\documentclass[journal]{IEEEtran}
\usepackage{graphicx}
\usepackage{comment}
\usepackage{amsmath,amssymb} 
\usepackage{color}
\usepackage{algorithm}
\usepackage{algorithmic}
\usepackage{epsfig}
\usepackage{graphics}
\usepackage{ulem}
\usepackage{color}
\usepackage{mathrsfs}
\usepackage[table]{xcolor}
\usepackage{booktabs}
\usepackage{makecell}
\usepackage{colortbl}
\usepackage{multirow}
\usepackage{array}
\usepackage{colortbl}
\usepackage{bm}
\usepackage{balance}
\usepackage[switch,columnwise]{lineno}
\IEEEpubid{\begin{minipage}{\textwidth}\ \\[12pt]  \centering \\
  \vspace{1.0cm}
  This article has been accepted for publication in a future issue of this journal, but has not been fully edited.  Content may change prior to final publication. \\ Citation information: DOI 10.1109/TCSVT.2021.3083257, IEEE Transactions on Circuits and Systems for Video Technology. \\
  1051-8215 \copyright 2021 IEEE. Personal use is permitted, but republication/redistribution requires IEEE permission.\\
  See http://www.ieee.org/publications standards/publications/rights/index.html for more information.
\end{minipage}}
\ifCLASSINFOpdf
\else
\fi
\begin{document}
\title{Image-to-Video Generation via 3D Facial Dynamics}

%
\author{Xiaoguang Tu,
        Yingtian Zou,
        Jian Zhao,~\IEEEmembership{Member, ~IEEE,
        Wenjie Ai,
        Jian Dong$^*$,
        Yuan Yao, \\
        Zhikang Wang,
        Guodong Guo,
        Zhifeng Li,
        Wei Liu, ~\IEEEmembership{Senior Member, ~IEEE} and
        Jiashi Feng,~\IEEEmembership{Member},~IEEE}
\thanks{Xiaoguang Tu is with Aviation Engineering Institute, Civil Aviation Flight University of China, Guanghan, China. Email: xguangtu@outlook.com.}%
\thanks{Wenjie Ai is with School of Information and Communication Engineering, University of Electronic Science and Technology of China, Chengdu, China. 201821011405@std.uestc.edu.cn.}%
\thanks{Yingtian Zou, and Jiashi Feng are with National University of Singapore, Singapore. Email: \{zouyingt@comp.nus.edu.sg, elefjia@nus.edu.sg.\}}%
\thanks{Yuan Yao is with Pensees Pte Ltd, Singapore. Email: distantyy@gmail.com.}%
\thanks{Zhikang Wang is with School of Electronic Engineering, Xidian University. Email: zkwang00@gmail.com.}%
\thanks{Jian Dong is with Shopee, Singapore. Email: jian.dong@shopee.com.}%
\thanks{Jian Zhao is with the Institute of North Electronic Equipment, Beijing, China. Homepage: https://zhaoj9014.github.io/. Email: zhaojian90@u.nus.edu.}%
\thanks{Guodong Guo is with  Institute of Deep Learning, Baidu Research, Beijing, China. Email: guodong.guo@mail.wvu.edu.}%
\thanks{Zhifeng Li and Wei Liu are with Tencent AI Lab, Shenzhen, China. Email: \{michaelzfli@tencent.com, wl2223@columbia.edu.\}}%
\thanks{$^*$ Jian Dong is the corresponding author.}%
\thanks{Manuscript received ; revised August .}}

\markboth{Journal of IEEE Transactions on Circuits and Systems for Video Technology, December~2020}%
{Shell \MakeLowercase{\textit{et al.}}: Bare Demo of IEEEtran.cls for IEEE Journals}
\maketitle

\begin{abstract}
We present a versatile model, FaceAnime, for various video generation tasks from still images. Video generation from a single face image is an interesting problem and usually tackled by utilizing Generative Adversarial Networks (GANs) to integrate information from the input face image and a sequence of sparse facial landmarks. However, the generated face images usually suffer from quality loss, image distortion, identity change, and expression mismatching due to the weak representation capacity of the  facial landmarks. In this paper, we propose to ``imagine'' a face video from a single face image according to the reconstructed 3D face dynamics, aiming to generate a realistic and identity-preserving face video, with precisely predicted pose and facial expression. The 3D dynamics reveal changes of the facial expression and motion, and can serve as a strong prior knowledge for guiding highly realistic face video generation. In particular, we explore face video prediction and exploit a well-designed 3D dynamic prediction network to predict a 3D dynamic sequence for a single face image. The 3D dynamics are then further rendered by the sparse texture mapping algorithm to recover structural details and sparse textures for generating face frames. Our model is versatile for various AR/VR and entertainment applications, such as face video retargeting and face video prediction. Superior experimental results have well demonstrated its effectiveness in generating high-fidelity, identity-preserving, and visually pleasant face video clips from a single source face image.
\end{abstract}

%
\IEEEpeerreviewmaketitle

\section{Introduction}
\IEEEPARstart{A}{s} an interesting topic in computer vision, face generation can be viewed as a task of image-to-image generation conditioned on a input face image. To address this issue, conditional Generative Adversarial Networks (cGAN) \cite{mirza2014conditional,nguyen2017shadow,songsri2019face,gauthier2014conditional,antipov2017face,kong2019cross} are usually deployed, taking additional conditions such as facial landmarks as input, to generate face images with specific poses or expressions
\cite{song2018geometry,hu2018pose,zhao2017dual,zhao20183d,zakharov2019few,zhao2020recognizing}.

\begin{figure}[!t]
\hspace{-0.0cm}
\centering{\includegraphics[width = 9.0cm, height=10.5cm]{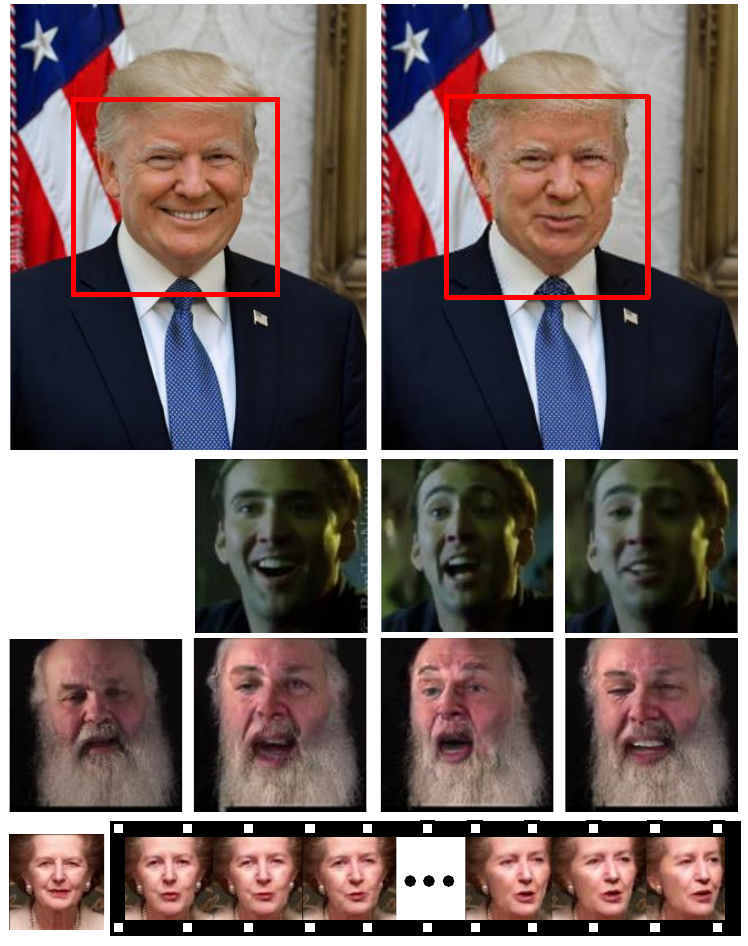}}
\vspace{-0.8cm}
\caption{Various interesting applications of our FaceAnime. From top to the bottom are facial expression manipulation, face retargeting and face video imagination from a single face image, respectively.
\label{fig:0}}
\end{figure}

Recently, this task has evolved to be a more challenging image-to-video translation problem  
which takes as input a single face image and generates specific face video clips. Many interesting applications have emerged based on this new technique such as facial expression manipulation, face retargeting and face video imagination from a single face image, as shown in Fig.~\ref{fig:0}.
Current methods~\cite{fan2019controllable,songsri2019face,tu2017illumination2,tu2017illumination} decompose a face image to a \textbf{content} part and a \textbf{motion} part, with the texture of source face image taken as content while motion realized as a stochastically dynamic process based on a Recurrent Neural Network (RNN) \cite{mikolov2011extensions,gregor2015draw,du2015hierarchical} to predict the temporal dynamics for video generation.
In \cite{fan2019controllable}, Fan \textit{et al.} address face video generation by predicting sparse facial landmarks from a single face image, but they only focus on facial expression generation with facial poses fixed during the generation process.
Later, \cite{songsri2019face} considers a more complicated case where both expressions and poses are changing during face sequence generation. \cite{songsri2019face} also uses sparse facial landmarks to represent dynamic changes across video frames. Though effective, their synthesized faces suffer from quality loss when the predicted pose has a wide variance. In a very recent work \cite{zakharov2019few}, a new framework is proposed to create a talking head from a handful of photographs by few-shot learning. Instead of using only one face image as the source, the method needs a few more photographs to increase the fidelity of personalization since the results from a single face image are not very ideal.
Similar to \cite{fan2019controllable} and \cite{songsri2019face}, this method also uses sparse facial landmarks to guide face generation.

Although sparse facial landmarks \cite{song2018geometry,hu2018pose,zakharov2019few,zhao2020recognizing,tu2019enhance,tu2020learning} and other 2D knowledge such as face/human skeleton \cite{yu2020multimodal,ma2021spatial,zhao2019multi,zhao2018towards,2020Towards,2019Look,2017Marginalized}, body boundary \cite{xu2020fine}, and motion residual \cite{lin2020motion,tu2017automatic} are somewhat viewed as a natural way to represent the dynamics of facial motions in previous literatures, we argue that they are defective for dynamic representation according to the following considerations:
\begin{itemize}
\item The sparse 2D landmarks cannot well represent the geometry of a face image. The global face shape and facial structural details are absent, which may lead to image distortion and quality loss for the synthesized images.

\item The sparse 2D landmarks do not carry any content information of the source  face image, which may cause the generated images overfit to the facial appearances only contained the training set.

\item Face identity should be preserved during video generation, however the sparse 2D landmarks have no identity information, bringing in identity changes for the synthesized results.
\end{itemize}

In view of these drawbacks, in this work we propose a novel method to generate face videos from 3D face dynamics.
A natural idea to make use of the 3D information is to replace the sparse 2D facial landmarks with the dense 3D facial landmarks. However, dense 3D facial landmarks contain thousands of 3D coordinates which are ill-suited to be predicted via LSTMs \cite{hochreiter1997long,jia2015guiding,zhang2016video}. Moreover, the dense 3D points do not include any content from the source image.
To solve these issues, we model the task of motion generation by predicting 3D dynamics with the 3D Morphable Model (3DMM) \cite{blanz1999morphable,zhu2016face,tu20203d,zhao20183d2,tu2021joint}. The 3D dynamics contain sub-components for facial expression, pose and global face shape, respectively, and can be easily predicted by LSTMs with low-dimensional vectors. Moreover, we further develop a sparse texture mapping algorithm to render the 3D dynamics, the obtained results contain shape, content and structural details that can be better utilized to guide face generation.

Our framework, which is termed as FaceAnime, contains two stages. In the first stage, a face image is fed into a 3D Dynamic Prediction (3DDP) network for predicting a 3D dynamic sequence based on the 3DMM.
The second stage is image generation conditioned on the predicted 3D dynamics.
Different from sparse 2D facial landmarks, the predicted 3D dynamics are rendered to involve rich information of face global shape, structure details and image content, which leads to more realistic faces.
As 3DMM coefficients contain independent components for face shape, expression and pose, it can generate face frames regarding part or all of the components. In this sense, the proposed method is versatile and applicable for various face video generation scenarios. In this paper, we focus on three image-to-video translation tasks, \textit{i.e.,} face video retargeting (prediction network is not used), video prediction and target-driven video prediction, respectively. Given a source face image, FaceAnime is able to retarget the poses and expressions of a reference video to the source face (\textbf{face video retargeting}). FaceAnime can also predict a stochastic and reasonable future for the source face (\textbf{video prediction}) or control the video generation towards a target face (\textbf{target-driven video prediction}). Qualitative and quantitative experimental results have demonstrated the effectiveness of our model by comparing with the other state-of-the-art face generation method. The source code and pretrained models will be available upon the acceptance of our paper.

Our contributions are summarized as follows:
\begin{itemize}
\item We propose a novel framework for highly-realistic and identity-preserving face video generation from a single face image, by exploiting the  informative 3D facial dynamics instead of the widely used 2D sparse facial landmarks.

\item We design a 3D Dynamic Prediction (3DDP) network to predict a spatio-temporal consistent 3D dynamic sequence.

\item We propose a sparse texture mapping algorithm to render the 3D face dynamics and use them as prior knowledge, which covers sparse content of the source image for guiding face generation.

\item We investigate face video prediction in both stochastic and controllable manners. The former predicts reasonable future face frames, while the latter ``imagines'' face frames towards a given target.


\end{itemize}

\subsection{Related Works}
We review the related literature on static and temporal face generation.


\subsubsection{Static Face Generation}
This task is typically tackled based on face morphable models \cite{amberg2009weight,cao2013facewarehouse,yang2011expression} by editing the geometry of a reconstructed face shape. With the rise of generative modeling, a number of methods apply Variational Auto-Encoder (VAE) \cite{shu2017neural,shu2018deforming,bao2017cvae} and Generative Adversarial Network (GAN) \cite{choi2018stargan,pumarola2018ganimation,wu2018reenactgan,qian2019make} to generate face images towards specific attributes. For example, some works \cite{bao2017cvae,shu2018deforming} use VAE and its variants to explore facial property disentanglement in the latent space.
Neural Face Editing \cite{shu2017neural} and Deforming Autoencoders \cite{shu2018deforming} utilize graphics rendering context such as UV maps, albedo, and shading to disentangle the latent representations.
However, such VAE-based methods may get blurry results due to element-wise divergence measurement and imperfect network architectures.
In \cite{bao2017cvae}, Bao \textit{et al.} combine VAE and GAN to a unified framework with an asymmetric training loss and fine-grained category labeling.
The starGAN \cite{choi2018stargan} exploits cycle consistency to change key attributes, like facial expressions, between the source and target images.
GANimation \cite{pumarola2018ganimation} takes a step further, translating a facial image according to the activation of certain facial Action Units (AUs\footnote{AUs is a system of taxonomy for classifying motions of human facial muscles.}) \cite{lucey2010automatically} and their intensities. However, the AUs cannot describe all possible localized motions such as lip motion patterns. As a result, GANimation cannot be used in a straightforward manner for realistic face generation.
In \cite{shen2018faceid}, Shen \textit{et al.} achieve identity-preserving face manipulation by FaceID-GAN, which uses a classifier of face identity as a third player, and makes it compete against the generator for distinguishing identities of real and synthesized faces.
In \cite{hu2018pose}, Hu \textit{et al.} propose a CAPG-GAN by training a couple-agent discriminator to generate face images towards different head poses.

\subsubsection{Temporal Face Generation}
There have been several previous works \cite{bansal2018recycle,wiles2018x2face,huang2017dyadgan} addressing face video generation.
RecycleGAN \cite{bansal2018recycle} translates video content from one domain to another while preserving the style native to the target domain.
In addition to cyclic consistency loss, it imposes spatio-temporal constraints over spatial constraints for effective face retargeting.
The X2Face \cite{wiles2018x2face} proposes to drive face synthesis using codes produced by embedding networks. Based on driving videos or other conditions such as head poses or audio inputs, the method generates a sequence of driving codes which in turn move the embedded face image for producing a target video.
The DyadGAN \cite{huang2017dyadgan} modifies facial expressions based on the sketch of sparse facial landmarks. The framework consists of two stages, one of which estimates sketched images of the target domains from the source video, and the other generates face images based on the estimated sketch.

\begin{figure}[!t]
\hspace{-0.0cm}
\centering{\includegraphics[width = 8.5cm, height=6cm]{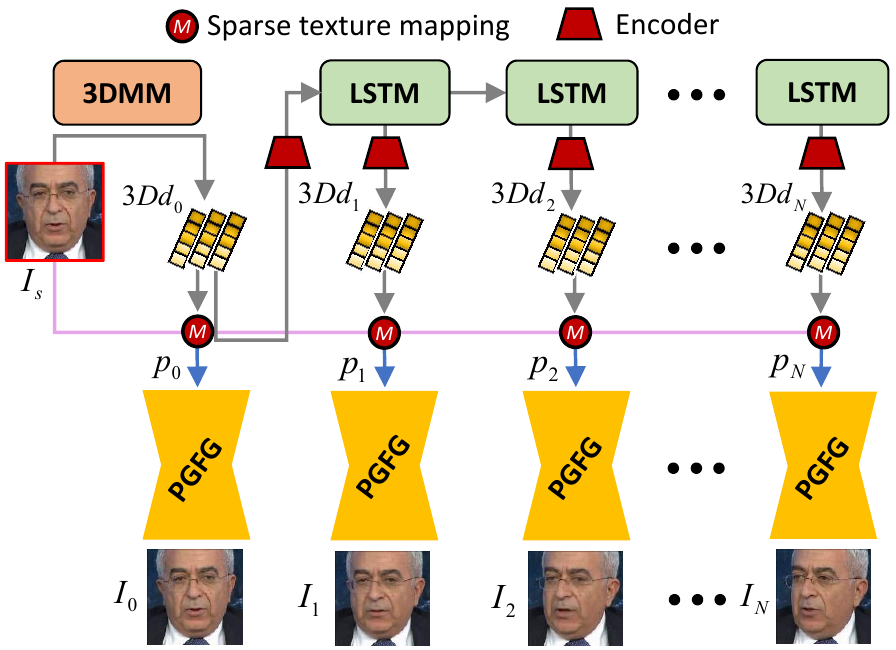}}
\vspace{-0.3cm}
\caption{The overview of FaceAnime. The future 3D dynamic  sequence ($3Dd_i$) of a source face is predicted by the 3DDP network. The source face along with the generated $3Dd_i$ is used to generate the texture prior ($p_i$), which is used to guide face generation.
\label{fig:1}}
\end{figure}

\begin{figure*}[t!]
\centering{\includegraphics[width = 18.0cm, height=7cm]{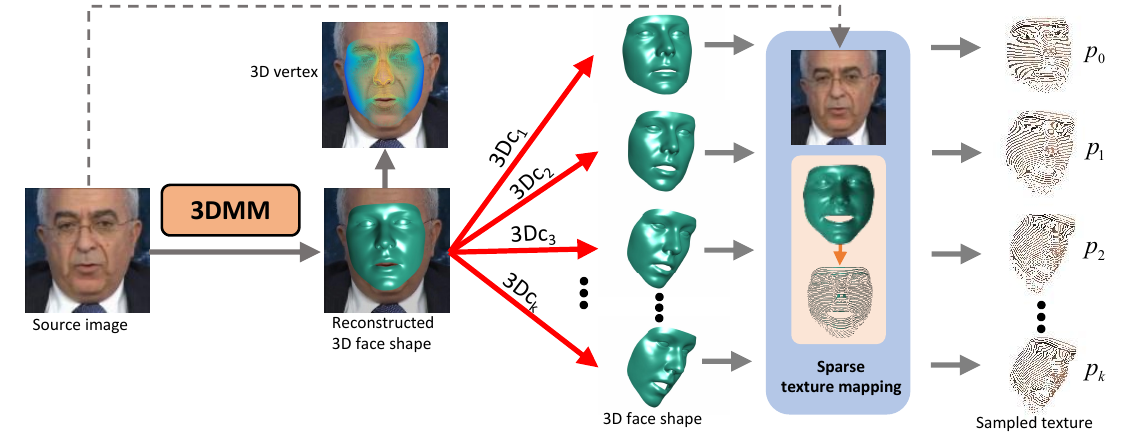}}
\vspace{-0.3cm}
\caption{The illustration of 3D face reconstruction and sparse texture mapping w.r.t. different 3DMM coefficients. Given a face image, the 3D face shape and vertices can be obtained by estimating the coefficients of a 3DMM. By changing the 3DMM coefficients to $3Dc_1, 3Dc_2$, ..., $ 3Dc_k$ , the 3D face can be re-shaped and its surface can be rendered based on the proposed sparse texture mapping. The sparsely rendered textures ($p_0, ..., p_k$) are used for guiding face generation. We use the 3DMM coefficients to represent the 3D dynamics $3Dd_1, 3Dd_2$, ..., $3Dd_k$.
\label{fig:2}}
\end{figure*}

Most video generation methods focus on temporal face retargeting from a given source face video, and few are devoted to predicting the future of a given face image. Compared with  human pose prediction \cite{chao2017forecasting,walker2017pose,tulyakov2018mocogan,chiu2019action} that is widely studied, face prediction is still under-explored which requires to recover rich facial details and realistic content instead of the coarse texture of human poses.
Two most related works to our approach are \cite{fan2019controllable} and \cite{songsri2019face}.
In \cite{fan2019controllable}, Fan \textit{et al.} tackle face prediction by predicting sparse facial landmarks from a source face. This method generates video clips of various lengths from a single face image according to a discrete facial expression label, such as happy, angry, surprise, \textit{etc}. \cite{songsri2019face} performs the  ``real'' future face generation. It predicts future frames by a Long Short-Term Memory network (LSTM) without any specified attributes, and generates reasonable temporal face frames only from a given face image.
In contrast, we focus on three face generation tasks, \textit{i.e.,} face video retargeting that retargets the pose and expressions of a reference video to the source face, video prediction which predicts reasonable future frames from a single face image, and target-driven video prediction that imagines possibly temporal changes between a given source face image and a target face image.

\section{Proposed Method}
Our FaceAnime contains a 3D Dynamic Prediction (3DDP) network and a Prior-Guided Face Generation (PGFG) network, as shown in Figure~\ref{fig:1}. The 3DDP network is used to predict future 3D dynamic sequences based on 3D Face Fitting. Such dynamics are then rendered by sparse texture mapping to guide face generation by the PGFG network.

\subsection{3D Face Fitting and Sparse Texture Mapping}
\subsubsection{3D Face Fitting}
3D face fitting is to estimate the 3D face shape, expression and camera projection matrix from the input 2D face image. Following \cite{tu20203d}, we estimate the 3D face from a single face image using the 3D Morphable Model (3DMM) \cite{zhu2016face,tu20203d}. The 3D vertices for describing a 3D face are estimated  from a 2D face and expressed as a linear combination over a set of   bases:
\begin{equation}
\begin{aligned}
\bm{S} =  \overline{\bm{S}} + \bm{A}_{\text{s}}\bm{\alpha}_{\text{s}} + \bm{A}_{\text{exp}}\bm{\alpha}_{\text{exp}},
\end{aligned}
\end{equation}
where $\overline{\bm{S}} \in \mathbb{R}^{3N}$ is the mean shape, $\bm{A}_\text{s} \in \mathbb{R}^{3N \times {40}}$ is the shape basis {trained on the 3D face scans}, $\bm{\alpha}_\text{s} \in \mathbb{R}^{40 \times {1}}$ is the shape representation coefficient, $\bm{A}_{\text{exp}} \in \mathbb{R}^{3N \times {10}}$ is the expression principle basis, $\bm{\alpha}_{\text{exp}} \in \mathbb{R}^{10 \times {1}}$ is the corresponding expression coefficient, and $N$ is the number of vertices.

The 3D face vertices $\bm{S}$ can be projected onto a 2D image plane with scale orthographic projection to generate a 2D face from a specified viewpoint:
\begin{equation}
\begin{aligned}
\bm{V} =  f*\text{Pr}*\bm{\Pi}* \bm{S} + t,
\end{aligned}
\end{equation}
where $\bm{V}$ represents the 2D coordinates of the 3D vertices projected onto the 2D plane, $f$ is the scale factor, $\bm{\Pi}$ is a fixed orthographic projection matrix, $\text{Pr}$ is the rotation matrix, and $t$ is the translation vector.
By minimizing the $\ell_{2}$ distance between projected landmarks and detected landmarks (inferred with the reconstructed 3D face model), the 3DMM coefficients $ [\text{Pr}, f, t, \bm{\alpha}_{\text{s}}, \bm{\alpha}_{\text{exp}}]$ can be estimated.

\subsubsection{Sparse Texture Mapping}
The 3D shape and vertices of a face image can be obtained based on the estimated 3DMM coefficients. As shown in Figure~\ref{fig:2}, giving a source face image, the 3D face shape of the source face can be modified arbitrarily by changing the reconstructed 3DMM coefficients. Then, the target sparse texture can be obtained by the modified 3DMM coefficients (3Dc$_1$, 3Dc$_2$, ... 3Dc$_k$), rendering by the source face image. In our face retargeting task, the modified 3DMM coefficients are obtained from the reference face video frames, while for the face prediction task the modified 3DMM coefficients are predicted by the LSTM modules.

However, the modified face shape does not carry any content from the source face, which may cause the generated face to overfit to the appearance of training images. Therefore, we perform texture mapping to render the reconstructed face shape by the source face image, which offers skin prior for face generation. We argue that using dense texture as prior is too strong, which may lead to undesired results in terms of subtle changes of the target motion. To address this issue, we perform sample interval during texture mapping, which we named sparse texture mapping, to impair the ability of dense prior, making the generated face better adapted to different motion changes. For example, if the eyes of the source face are closed but are desired to be open, when using dense texture to guide face generation, the closed eyes from dense texture serving as prior knowledge are so strong that the generated face can hardly show naturally open eyes.

\begin{figure*}[!t]
\hspace{-0.0cm}
\centering{\includegraphics[width = 18.5cm, height=6.8cm]{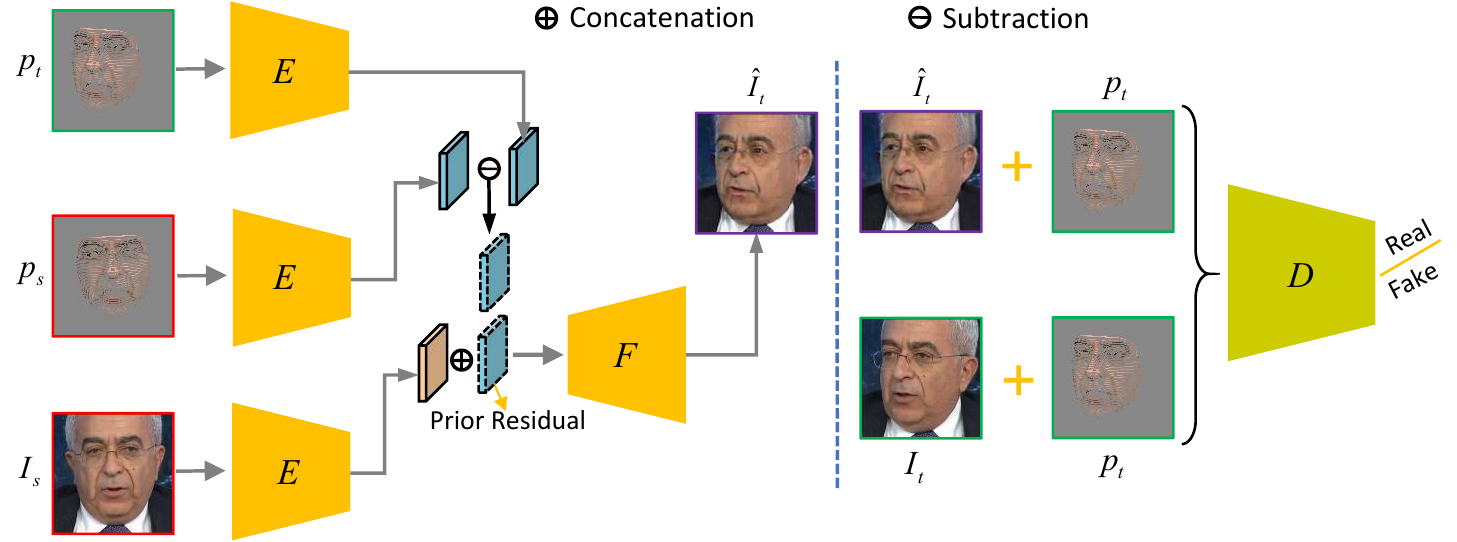}}
\vspace{-0.7cm}
\caption{The architecture of the PGFG network. PGFG contains a prior-guided face generator (left) and a dual input discriminator (right). The prior-guided generator contains an encoder $E$ and a decoder $F$. $F$ takes as input the concatenated representation of the input face image and the prior residual between the source and target priors. $D$ takes the target prior as the condition and pairs the real/fake images as the dual input.
\label{fig:1x}}
\end{figure*}

\begin{figure}[!t]
\hspace{-0.0cm}
\centering{\includegraphics[width = 8.5cm, height=3.0cm]{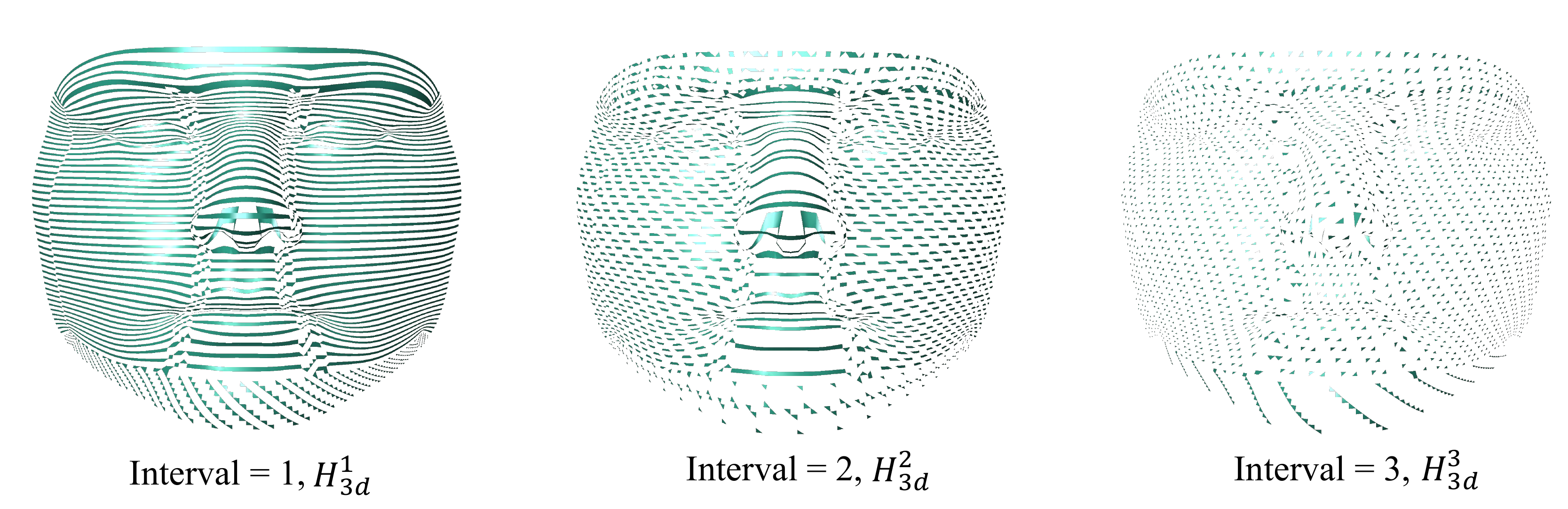}}
\vspace{-0.3cm}
\caption{The sparse mapping results with different interval values.
\label{fig:minor1}}
\end{figure}

Formally, a 3D face shape can be represented by a vertex matrix $\bm{V} \in \mathbb{R}^{3 \times {N}}$ and a 3D triangulated mesh $\bm{H} \in \mathbb{R}^{3 \times {K}}$. $\bm{V}$ stores the 3D coordinates of each vertex and $\bm{H}$ is a triple adjacency matrix, indicating the adjacency relation of triple vertices as
\begin{equation}
\begin{aligned}
\bm{H} =
\begin{bmatrix}
   a_1 & a_2 & ... & a_K \\
   b_1 & b_2 & ... & b_K \\
   c_1 & c_2 & ... & c_K
  \end{bmatrix},
\end{aligned}
\end{equation}
where $a_i, b_i$ and $c_i$ are the vertex indexes. For example, $\{a_i = 1, b_i = 3, c_i=100\}$ means that the $1^{st}$, $3^{rd}$ and $100^{th}$ vertices are connected for a triangle.
We define the 3D sparse texture mapping by
\begin{equation}
\begin{aligned}
\bm{H}^{n}_{3d}  = S_{3d}(\bm{H}^n_{2d}) = S_{3d}(\begin{bmatrix}a_1 & a_{2*n+1} & ... & a_{k*n+1}\\ b_1 & b_{2*n+1} & ... & b_{k*n+1}\\ c_1 & c_{2*n+1} & ... & c_{k*n+1}\\ \end{bmatrix}), \\
k = [\frac{K-1}{n}],
\end{aligned}
\end{equation}
where $\bm{H}^n_{2d}$ is the 2D down-sampled result at an interval of $n$ ($n$ is positive integer); $S_{3d}$ is the sparse mapping operation, where only the triple adjacencies that each element belongs to $\bm{H}^n_{2d}$ are reserved, while other adjacencies are removed; $[\cdot]$ is the round down operation. After performing the sparse mapping, the source face texture can be mapped to the target to obtain the sparse texture. Figure~\ref{fig:minor1} shows the sparse texture mapping results with different interval values. Our sparse texture mapping makes the target prior capable of carrying content from the source face, and meanwhile ensures the prior not to be too strong, in order to improve the adaptation ability of the model itself.

\subsection{3D Dynamic Prediction}
We leverage the 3D Dynamic Prediction (3DDP) network to predict 3DMM dynamics, which can be used for 3D face reconstruction and texture mapping to provide valuable information for face generation. Different with the method \cite{villegas2017learning} that only learns to predict a reasonable future from the initial state, we focus on three tasks, \textit{i.e.,} face video retargeting, video prediction the same as \cite{villegas2017learning} and target-driven video prediction that ``imagines'' the temporal changes via completing missing frames between the given source and target face images. For the retargeting task, we use a reference video to offer the sequential changes, rather than using 3DDP to predict the future changes.

\subsubsection{Video Prediction}
Given the observed dynamics (3DMM coefficients), the LSTM encodes them as
\begin{equation}
\begin{aligned}
& \bm{h}_{t-1} = \text{Enc}(\bm{d}_{t-1}), \\
& \left[\bm{h}_t, \bm{c}_t\right] = \text{LSTM}(\bm{h}_{t-1}, \bm{c}_{t-1}),
\end{aligned}
\end{equation}

where $\text{Enc}$ is an encoder, ${d_{t-1}}$ represents the observed dynamics up to time $t-1$, and ${\bm{c}_t}$ is the \textit{memory cell} that preserves information from the history of inputs. In order to predict a reasonable future motion, LSTM has to firstly observe a few motion inputs to identify the type of motion occurring in the pose sequence and how it is changing over time.

During training, the future dynamics can be generated by
\begin{equation}
\begin{aligned}
\hat{\bm{d}_t} = r(\bm{w^{\top}}\bm{h}_t) ,
\end{aligned}
\end{equation}
where $\bm{w}$ is a projection matrix, $r$ is a function on the projection (\textit{i.e.} tanh), and $\hat{\bm{d}_t} \in \mathbb{R}^{62}$ is the predicted 3DMM coefficient, representing the 3D dynamic at time $t$. Based on Eqn. (6), the model could learn a reasonable future from an initial dynamic $\bm{d}_0$. After obtained the facial motion sequence for the future frames, we proceed to generate the pixel-level visual future.

\subsubsection{Target-driven Video Prediction}
For a deep network to achieve target-driven video generation under various lengths, i) the model should be aware of the information of controlling images and ii) the model should be able to perceive time lapse and generate the targeted end-frame at the designated timestep

For a LSTM to achieve target-driven motion generation, the model needs to take two inputs, \textit{i.e.}, the source dynamic and the target dynamic.
The model should be aware of the information of controlling source/inital inputs and learns a dynamic sequence that changes from the source dynamic to the target. Different from future prediction, we use a time counter to re-weight the target dynamic, representing them as the predicted history. We formulate the LSTM prediction by
\begin{equation}
\begin{aligned}
& \bm{h}_{t-1} = \text{Enc}(\bm{d}_{t-1}), \quad \bm{h}_{T} = \text{Enc}(\bm{d}_{T}), \\
& \left[\bm{h}_t, \bm{c}_t\right] = \text{LSTM}(\bm{h}_{t-1}, \frac{t}{T}\bm{h}_T), \\
& \hat{\bm{d}_t} = r(\bm{w^{\top}}\bm{h}_t) ,
\end{aligned}
\end{equation}
where $\bm{d}_T$ represents the target dynamic, $T$ is the prediction length; $t = 0$ indicates the beginning of the sequence, and $t=T$ represents the end of the sequence. While the targeted dynamic is already fed as an initial frame, we randomly select one dynamic from each training sequence, which serves as a fixed point to strengthen the training process, preventing the prediction deviating from the desired predicted sequence.

\subsubsection{Loss Function}
Given a source face image, we use the 2DAL \cite{tu20203d} to regress its 3DMM coefficients, represented as the initial 3D dynamic $\bm{d}_0$. Our prediction model then observes $\bm{d}_0$ and proceeds to generate a coefficient sequence $\hat{\bm{d}}_{1:T}$. During training, we minimize two losses, \textit{i.e.}, the 3DMM coefficient loss and the 3D vertex loss. The 3DMM coefficient loss is defined by minimizing the Euclidean distance between the predicted 3DMM coefficients and corresponding ground-truth 3DMM coefficients:
\begin{equation}
\begin{aligned}
\mathcal{L}_{\text{3dc}} = \frac{1}{T}\sum_{t=1}^{T}||\hat{\bm{d}}_{1+t}  - \bm{d}_{1+t}||_2^2,
\end{aligned}
\end{equation}
where $\hat{\bm{d}}_{1+t}$ and $\bm{d}_{1+t}$ are the predicted and ground-truth 3DMM coefficients at time $1+t$, respectively.

However, the 3DMM coefficient is just an intermediate variable, the model has to learn a transformation from the coefficient to the reconstructed 3D shape. In addition, the 3DMM coefficient has no semantic meaning, while the 3D vertex contains rich semantic information, such as face shape, structure and depth. When using the combination of these two loss functions, the 3DMM coefficient serves as a coarse constraint to ensure the model fast converge to a reasonable result, while the 3D vertex constraint refines the reconstruction based on the semantic meaning, which may achieve better results. We design the 3D vertex loss by minimizing the Euclidean distance between the 3D vertices of the predicted 3DMM coefficients and corresponding 3D vertices of the ground-truth 3DMM coefficient:
\begin{equation}
\begin{aligned}
\mathcal{L}_{\text{3dv}} = \frac{1}{T}\sum_{t=1}^{T}||\hat{\bm{v}}_{1+t}  - \bm{v}_{1+t}||_2^2,
\end{aligned}
\end{equation}
where $\hat{\bm{v}}_{1+t}$ and $\bm{v}_{1+t}$ are the 3D vertices of predicted coefficients and ground-truth coefficients, respectively.

The overall loss function is a weighted sum of the 3DMM coefficient loss and the 3D vertex loss:
\begin{equation}
\begin{aligned}
\mathcal{L}_{\text{pred}} = \mathcal{L}_{\text{3dv}} + \lambda_1 \mathcal{L}_{\text{3dc}},
\end{aligned}
\end{equation}
where $\lambda_1$ is the weighting parameter controlling the weighting between different losses.

\subsection{Prior-Guided Face Generation}
The source face image is used to render the predicted 3D dynamics based on the proposed sparse texture mapping. These sparse textures are priors that we use for guiding face generation.

Our face generation model is a conditional GAN (cGAN), which contains a prior-guided face generator $G$ for high-fidelity face generation conditioned on the predicted and rendered prior sequences, and a discriminator $D$ to distinguish the real/fake of the outputs regarding the target priors. Figure~\ref{fig:1x} shows the architecture of our Prior-Guided Face Generation (PGFG) network.

\subsubsection{Architecture}
The generator $G$ takes in three inputs, including a source face image $I_s$, $I_s$'s corresponding texture prior $\bm{p}_s$ and a target texture prior $\bm{p}_t$. Instead of using $\bm{p}_t$ to directly guide face generation, we adopt the prior residual to guide generation. Specifically, the encoder accepts inputs $\bm{p}_s$ and $\bm{p}_t$. Then the motion residual is obtained by subtraction in the feature space $E(\bm{p}_t) - E(\bm{p}_s)$. We can get the target face by
\begin{equation}
\begin{aligned}
\hat{I}_t = G(I_s) = F(E(I_s) \small{\oplus}\lbrack E(\bm{p}_t) - E(\bm{p}_s) \rbrack),
\end{aligned}
\end{equation}
where $E$ and $F$ are the encoder and decoder of the prior-guided generator, respectively, and $\oplus$ represents the concatenation operation.

To enable the decoder to more effortlessly reuse features of different spatial positions and facilitate feature propagation, we add dense connections between the upsampling layers. The decoder has 6 dense blocks in total, the outputs of each block are connected to the first convolutional layers located in all subsequent blocks in the decoder.  As the blocks have different feature resolutions, we upsample feature maps with lower resolutions when we use them as inputs to higher resolution layers.

The discriminator has two inputs, \textit{i.e.}, the target face image's texture prior pairs with the target face image or the generated face image, $\lbrack \bm{p}_t, I_t \rbrack$ vs. $\lbrack \bm{p}_t, \hat{I}_t \rbrack$. The target texture prior $\bm{p}_t$ serves as a condition to criticize  the generated image $\hat{I}_t$, to judge whether it is consistent with $I_t$ w.r.t. the prior knowledge which is encoded with both facial dynamics and content information. Following such a condition, $G$ can generate face images that have the same content with the source face image $I_s$ but present diverse motions regarding the predicted 3D dynamics.

\begin{figure*}[!h]
\centering{\includegraphics[width = 18.0cm, height=12.2cm]{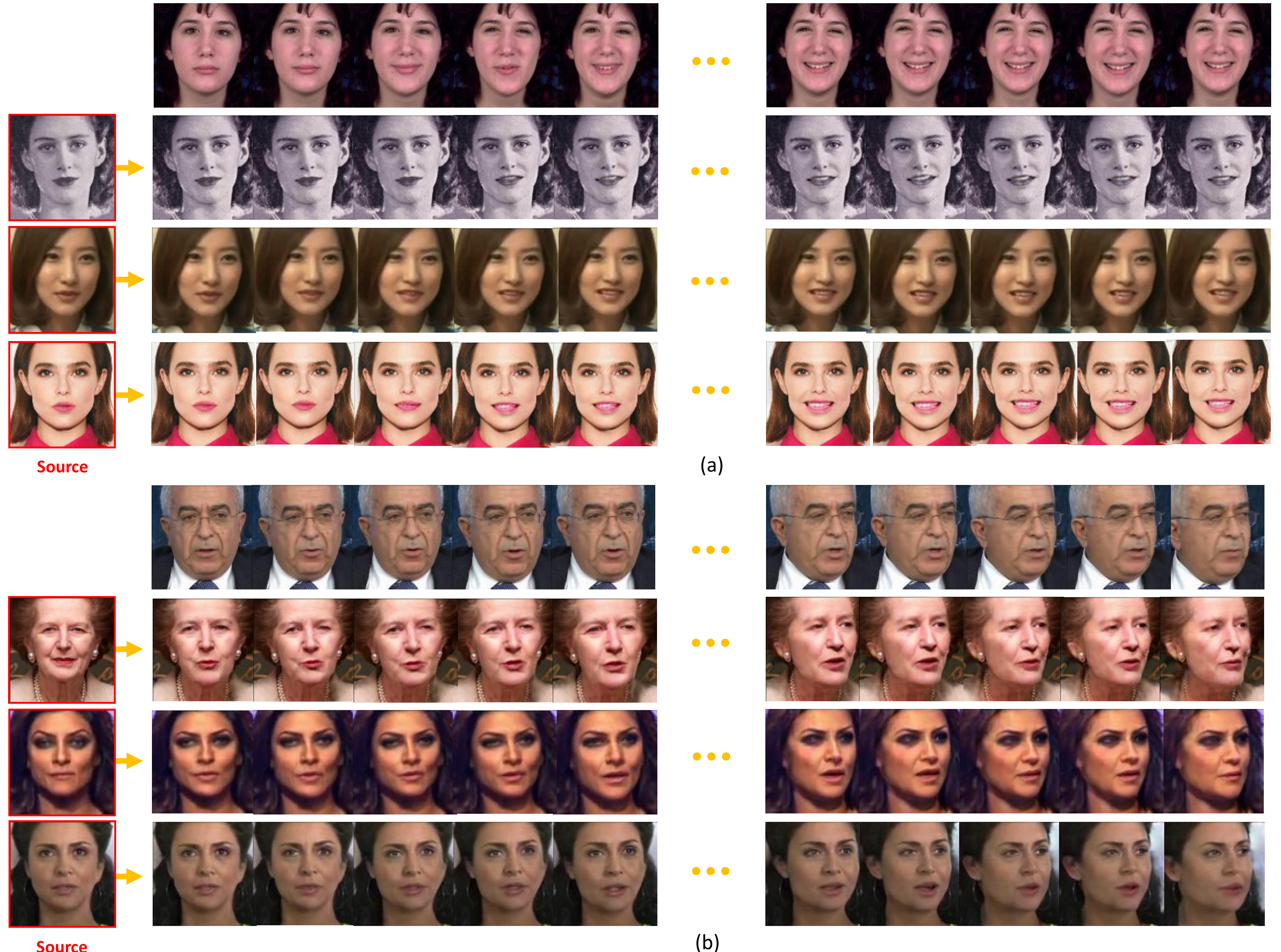}}
\vspace{-0.3cm}
\caption{Face video retargeting results. (a) Facial expression (smile) retargeting results. (b) Head talking retargeting results. The images of the first row are the reference frames selected from the MUG \cite{aifanti2010mug}  dataset. The images with red border are the source face images from the Small-Size dataset.
\label{fig:3}}
\end{figure*}

\subsubsection{Loss Function}
The process of PGFG network is supervised by three losses, \textit{i.e.,} image pixel-wise loss $\mathcal{L}_{\text{img}}$, conditional adversarial loss $\mathcal{L}_{\text{adv}}$ and identity-preserving loss $\mathcal{L}_{\text{id}}$. For $\mathcal{L}_{\text{img}}$, we penalize the pixel-wise Euclidean distance between the target face $I_t$ and the generated face $\hat{I}_t$ as
\begin{equation}
\begin{aligned}
\mathcal{L}_{\text{img}}(I_t, \hat{I}_t) = ||I_t - \hat{I}_t||_2^2.
\end{aligned}
\end{equation}

The dual-input discriminator is used to refine $\hat{I}_t$ conditioned on the target texture prior $\bm{p}_t$. The conditional adversarial loss can thus be defined as
\begin{equation}
\begin{aligned}
\mathcal{L}_{\text{adv}} =  \mathbb{E}_{I_s\in\mathcal{I}}[\log(D([\bm{p}_t, I_{t}]))+ \log(1-D([\bm{p}_t, G(I_s)]))].
\end{aligned}
\end{equation}

During training, the identity of generated face images should be preserved. To achieve this goal, we employ the identity-preserving loss $\mathcal{L}_{\text{id}}$, which draws the identity representations of $I_t$ and $\hat{I}_t$ closer in the normalized identity metric space. We use a pre-trained face recognition model $R_{\text{id}}$ to extract identity features and fix its parameters during training. $\mathcal{L}_{\text{id}}$ is defined as
\begin{equation}
\begin{aligned}
\mathcal{L}_{\text{\text{id}}}(I_t, \hat{I}_t) = \left\|\frac{R_{\text{id}}(I_t)}{||R_{\text{id}}(I_t)||_2} - \frac{R_{\text{id}}(\hat{I}_t)}{||R_{\text{id}}(\hat{I}_t)||_2}\right\|_2^2.
\end{aligned}
\end{equation}

The overall generation loss is a weighted sum of the above losses. The parameters of the generator ($\theta_{G}$) and those of the discriminator ($\theta_{D}$) are trained alternatingly to optimize the following min-max problem:
\begin{equation}\label{11}
\min\limits_{\theta_G} \max\limits_{\theta_{D}} \mathcal{L}_{\text{gen}} = \mathcal{L}_{\text{\text{img}}}(I_t, \hat{I}_t) + \lambda_2 \mathcal{L}_{\text{\text{id}}}(I_t, \hat{I}_t) + \lambda_3 \mathcal{L}_{\text{\text{adv}}},
\end{equation}
where $\lambda_2$ and $\lambda_3$ are the weight constants of the corresponding loss functions.

\section{Experiments}
In this section, we first verify the efficacy of FaceAnime on face video retargeting and face video prediction. Then, we evaluate the quality of generated face images by comparing with Pix2pixHD \cite{wang2018high}, X2Face \cite{wiles2018x2face} and FSGAN \cite{nirkin2019fsgan}. FSGAN is a two-stage framework that contains first face retargeting and second face swapping. In this case, we only use the first stage to train a face retargeting model and then compare with our FaceAnime. After that we provide our component analysis.

\subsubsection{Datasets}
1) We use the IJB-C \cite{maze2018iarpa} dataset to train FaceAnime. This dataset contains video frames and still images of 3,531 persons collected from the Web. It has no overlap with popular face recognition benchmarks. We select 2,000 high-quality videos from its video subset for training and 448 videos for evaluation, where each video has no more than 80 frames.
2) The CASIA-webFace \cite{yi2014learning} is used to train the face recognition model $R_{\text{id}}$. This dataset consists of 494, 414 images from 10,575 subjects.
3) The MUG~\cite{aifanti2010mug} dataset is used to evaluate facial expression retargeting. This dataset contains 86 subjects (35 female and 51 male) with six basic expressions, \textit{i.e.,} anger, disgust, fear, happiness, sadness and surprise. It is commonly used for the task of facial expression generation.
4) The 300VW \cite{zafeiriou2015300} dataset is used to evaluate the methods on the task of facial landmark detection.
5) We also select some celebrity face images from the Internet, which we call Small-Size dataset, to show the visual results of our method in the ``in-the-wild'' environment. The small-size dataset has no overlap with IJB-C.

\subsubsection{Implementation Details}
Throughout the experiments, the size of face images is fixed as $128 \times 128$; the constraint factors $\lambda_1$, $\lambda_2$, $\lambda_3$ are fixed as $1\times 10^{3}$, $1\times 10^{4}$, and 1, respectively; the batch size is set to 8; the prediction length for the 3DDP network is set as 24; the initial learning rates $lr$ for the generator and discriminator are $10^{-4}$  and 1, respectively. $lr$ decreases 10 times at the end of each epoch. We use the off-the-shelf 2DAL \cite{tu20203d} for 3DMM coefficients regression. We initialize the face recognition network with ResNet-50 \cite{he2016deep} as the backbone to extract identity features, which is pre-trained on the CASIA-Webface \cite{yi2014learning} dataset using the AAM loss \cite{deng2019arcface}.

\begin{figure*}[!t]
\centering{\includegraphics[width = 18.0cm, height=21cm]{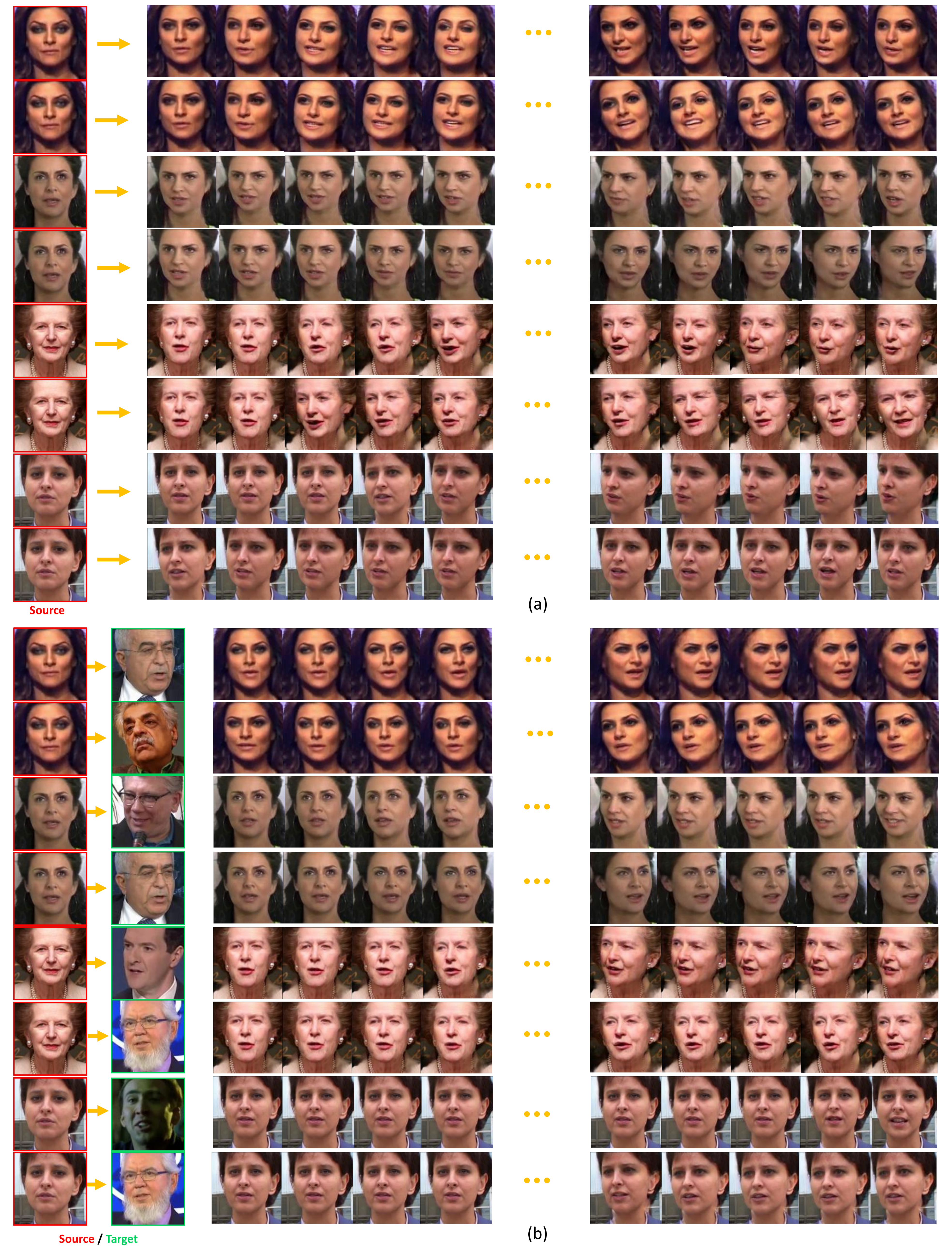}}
\vspace{-0.4cm}
\caption{Face video prediction results. (a) Video prediction results. (b) Target-driven video prediction results. The images with red border are the source face images, the images with green border are the target face images.
\label{fig:4}}
\end{figure*}

\subsection{Results of Face Video Generation}
We show video generation results of our FaceAnime on the tasks of face video retargeting, video prediction and target-driven video prediction, respectively.

For face video retargeting, we first perform facial expression retargeting, where only facial expressions of the generated face frames are changed while poses are fixed for each frame. To achieve this, we fix the pose and shape components of the reconstructed 3DMM coefficients, and only change the expression component during inference. The generation results are shown in Figure~\ref{fig:3} (a). We can see that FaceAnime is able to generate consecutive smiling face frames with high realism and impressive quality, transferring the smiling expressions of reference video frames to a single source face. We then perform a more challenging retargeting task, \textit{i.e.,} head talking retargeting, where both facial poses and expressions are changed following the given reference frames. The head talking retargeting results are shown in Figure~\ref{fig:3} (b). In this application, the shape component of 3DMM coefficients for a given face image is fixed, while pose and expression components of 3DMM coefficients are changed following the reference frames. As we can see, FaceAnime is able to retarget the desired poses and talking expressions extracted from the reference video to the source face.

For face video prediction, two sub-tasks are performed to show the effectiveness of FaceAnime, \textit{i.e.,} video prediction and target-driven face prediction. The same as head talk retargeting, the shape component of 3DMM coefficients for the source face image is fixed, and the pose and expression components of 3DMM coefficients are predicted by the 3DDP network during the inference. For each prediction  task, we randomly select one video and one single face image from IJB-C testing set as the reference video and the source face, respectively. For video prediction, we first use the 3D Dynamic Prediction (3DDP) network to predict a reasonable future motion sequence from the source face. The predicted motion sequence is used to guide face generation conditioned on the given source face. For target-driven prediction, we need two images as inputs: one as the source face and the other as the target face. The 3DDP network is used to predict a smooth motion changing from the source face to the target, which is subsequently used to guide face generation conditioned on the source face image. The generation results of the two sub-tasks are shown in Figure~\ref{fig:4} (a) and (b), respectively. As can be seen, the predicted face images are realistic, high-fidelity and identity-preserving across all possible pose and expression changes.

For the results of \textit{Rows} 1 \& 7 in Figure~\ref{fig:4} (b), the expressions are not very similar to the target. This might be the limitation of the 3D face reconstruction model that we have used. We follow the work 3DDFA \cite{zhu2016face} to obtain the 3D face vertices, to include face structural details and expression information for face generation. However, 3DDFA only use a 62-dimensional 3DMM coefficient to represent a 3D morphable model, only 10-dimensional coefficients are used for expression representation, which may be hardly to capture the subtle expression changes.

\begin{figure*}[!t]
\centering{\includegraphics[width = 17.5cm, height=14cm]{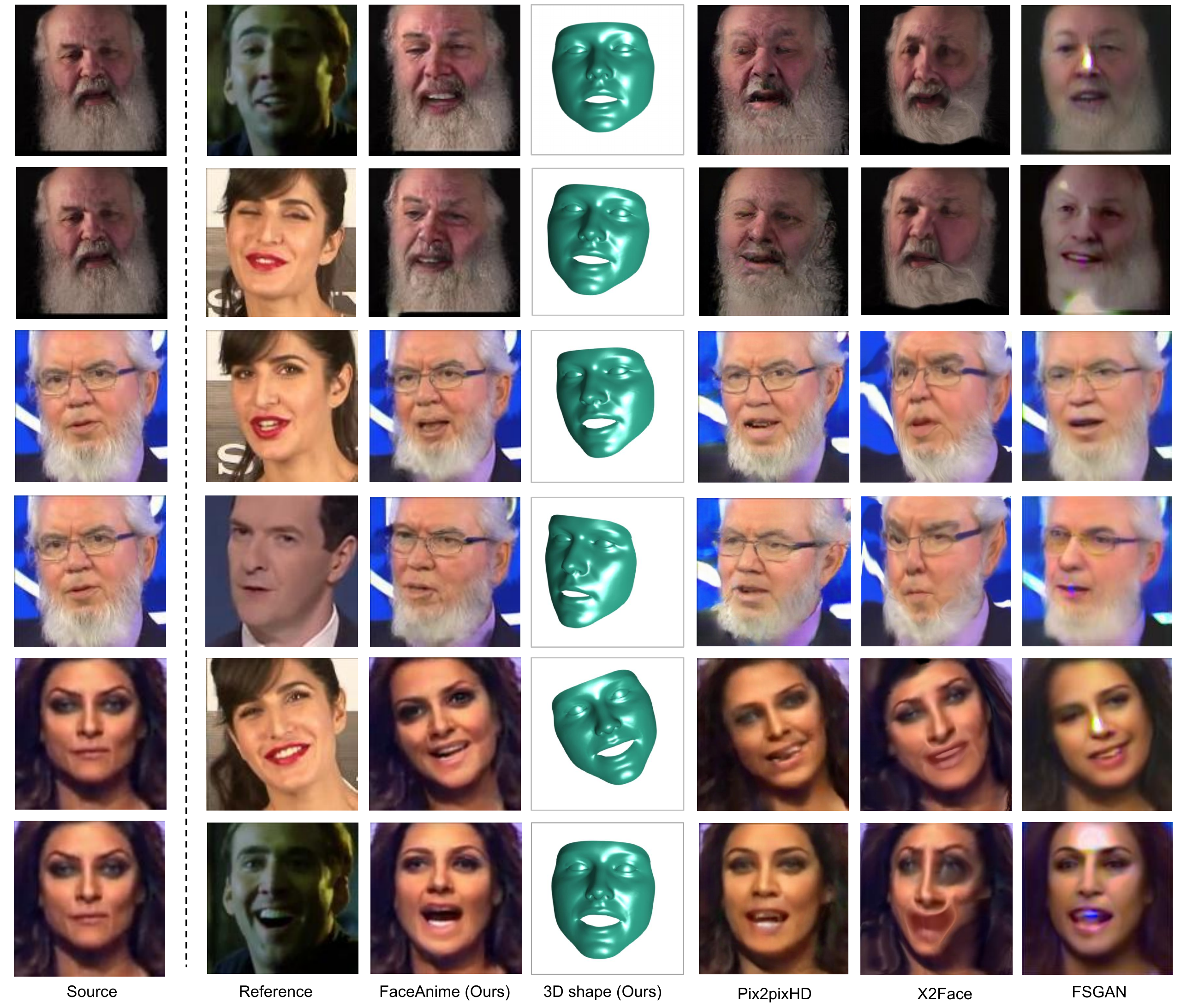}}
\vspace{-0.4cm}
\caption{Qualitative comparison results among FaceAnime, Pix2pixHD \cite{wang2018high}, X2Face \cite{wiles2018x2face} and FSGAN \cite{nirkin2019fsgan} on the IJB-C testing dataset. FaceAnime generates highly realistic face images that precisely follow the expressions and poses of the reference images. The results in \textit{Col.} 4 are the reconstructed 3D shape by FaceAnime.
\label{fig:5}}
\end{figure*}

\begin{figure*}[!t]
\centering{\includegraphics[width = 18.0cm, height=3.6cm]{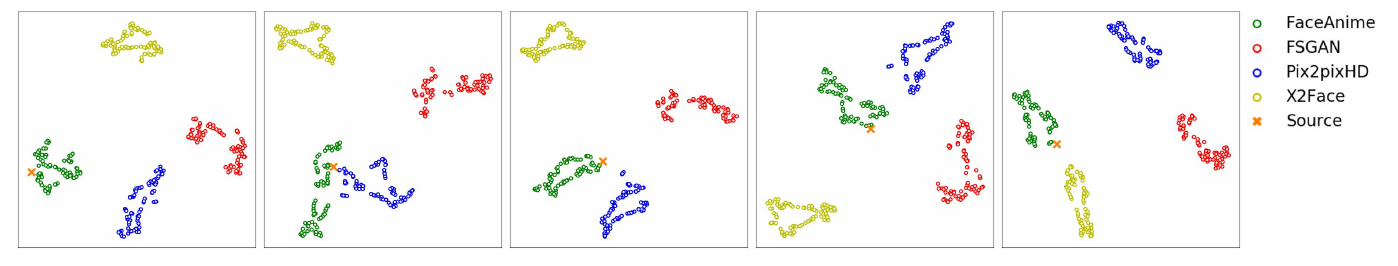}}
\vspace{-0.3cm}
\caption{The distribution of identity features corresponding to results from FaceAnime, Pix2pixHD, X2Face and FSGAN. Each sub-figure represents one identity. The cross mark denotes the feature of the source face for each identity; the circles plots denote the features of the generated faces. The identities are randomly selected from the IJB-C testing set. We use t-SNE \cite{van2014accelerating} to reduce the dimensions for figure plotting. (Best viewed in color)
\label{fig:6}}
\end{figure*}

\subsection{Comparison with State-of-the-Arts}
We quantitatively compare the results of our FaceAnime with Pix2pixHD \cite{wang2018high}, X2Face \cite{wiles2018x2face} and FSGAN \cite{nirkin2019fsgan} by computing the widely used image reconstruction metric Peak Signal-to-Noise Ratios (PSNR), Structural Similarity index (SSIM) and Frechet Inception distance (FID). For PSNR and SSIM, the higher value is the better, while for FID the lower value the better. For fair comparison, we randomly select 20 video clips from IJB-C testing set, where each video clip contains 40-80 consecutive frames of the same person. For each video clip, we select a single face image with good quality as the source face to perform face retargeting and then calculate PSNR, SSIM and FID. In addition, we use a state-of-the-art face alignment method \cite{bulat2017far} to locate landmarks of the generated results and compare them with the ground-truth landmarks on 300VW \cite{zafeiriou2015300}. We then measure the error between landmarks with the point-to-point Root Mean Square error. We call this evaluation metric Landmark Root Mean Square (LRMS).

\begin{table}[t!]
\small
\begin{center}
\caption{Quantitative comparison with other methods on the IJB-C testing set.}
\vspace{-0.2cm}
\resizebox{0.48 \textwidth}{!}{
\begin{tabular}{c l c c c c c}
\hline
1 & Methods \quad & PSNR $\uparrow$ & SSIM $\uparrow$ & FID $\downarrow$ & LRMS $\downarrow$   \\
\hline
2 & Pix2pixHD\cite{wang2018high} & 29.83 & 0.62 & 74.04 & 0.35 \\
3 & X2Face \cite{wiles2018x2face} & 29.61 & 0.61 & 97.70 & 0.46 \\
4 & FSGAN \cite{nirkin2019fsgan} & 29.12 & 0.58 & 53.21 & 0.27\\
5 & FaceAnime (Ours) & \textbf{29.87} & \textbf{0.65} & \textbf{49.10} & \textbf{0.21}\\
\hline
\end{tabular}}
\end{center}
\label{tab:1}
\end{table}

\begin{figure*}[!t]
\centering{\includegraphics[width = 18.0cm, height=6.8cm]{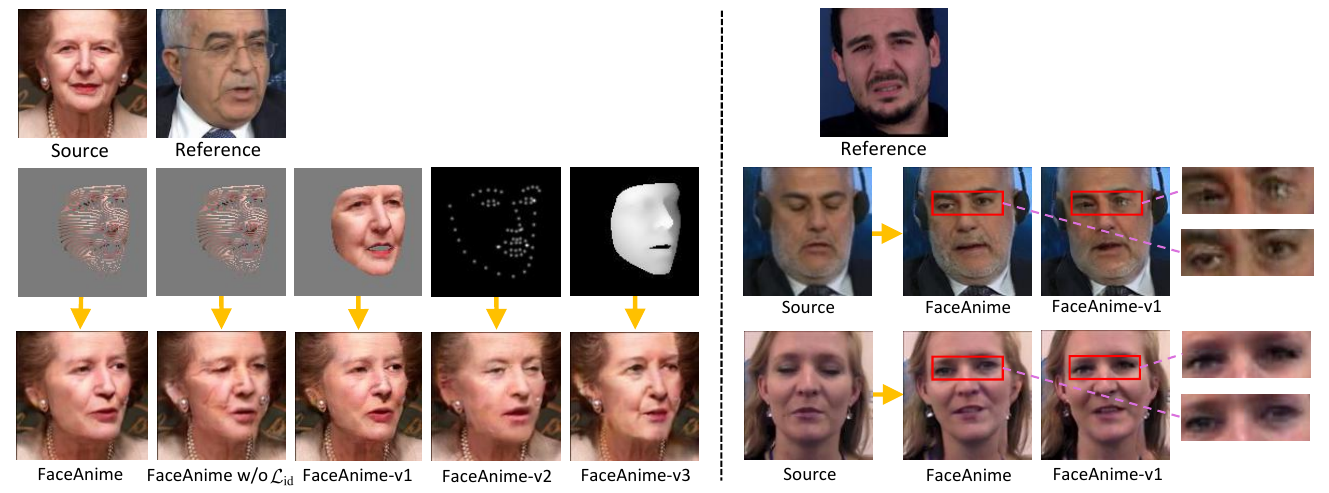}}
\vspace{-0.3cm}
\caption{Left: components analysis results. Right: comparison with dense texture as a prior in a special case (eyes of the source face are closed).
\label{fig:7}}
\end{figure*}

The results are reported in Table~\uppercase\expandafter{\romannumeral1}. As we can see, FaceAnime consistently outperforms other methods in terms of all evaluation metrics. First, we compare FaceAnime with Pix2PixHD and X2Face. As all the three methods use $\ell_2$ loss for optimization, they  achieve comparable PSNR and SSIM values (the SSIM and PSNR values of FaceAnime are slightly higher than the other two). However, for FID which mostly measures the perceptual realism of the generated face image, our FaceAnime performs much better than Pix2pixHD and X2Face. Even compared with Pix2pixHD that maximizes only the perceptual metric and thus gives high FID value, our FaceAnime is better by 24.94, which is a large margin. For FSGAN, it achieves the lowest PSNR and SSIM values, mainly due to the highlight region and over-smoothing of the results. FaceAnime achieves the lowest LRMS value, indicating that the poses and expressions of its generated face images match better with the target than the other methods.

To qualitatively compare our FaceAnime with other methods, we select 3 source face images and the other 3 reference images from IJB-C testing set to perform face retargeting. The visual comparison results are shown in Figure~\ref{fig:5}. As can be seen, our FaceAnime achieves much higher natural-looking and high-quality results than those by Pix2pixHD, X2Face and FSGAN. Pix2pixHD is able to generate face images with overall good quality. However, for challenging regions such as eyes and mouth, the results contain obvious artifacts. For X2Face, even though it well addresses such challenging regions, the output face is distorted for larger facial expression (\textit{Rows} 4 \& 6) and the results cannot well match the target pose and expression (see \textit{Row} 5 in Figure~\ref{fig:5}). FSGAN can generate results that well match the target pose and expression, however it brings in highlight regions into the generated image (see \textit{Rows} 1, 2, 5 $\&$ 6 in Figure~\ref{fig:5}), and the generated images are over-smoothed. Seen from \textit{Row} 5, the identities of the outputs of Pix2pixHD, X2Face and FSGAN are changed. In comparison, with the identity-preserving loss $\mathcal{L}_{\text{\text{id}}}$ used, our FaceAnime preserves the identity very well for generated face images. We also visualize the learned identity features in Figure~\ref{fig:6} to show the capability of FaceAnime in identity preservation. Specifically, we randomly select 1 video clip as the reference video and 5 source face images from the testing set. Then, we use the four methods to perform face retargeting tasks. Each method generates 5 video clips for the 5 source face images, respectively. We use an off-the-shelf face recognition model \cite{liu2017sphereface} to extract the identity features for each face image for visualization. As can be seen from Figure~\ref{fig:6}, for each face, the outputs from FaceAnime are scattered near the source face in the identity metric space, while for the other methods, the outputs are far away from the source face. The results show that our FaceAnime performs significantly better than the other two methods on face identity preservation, regardless of facial pose and expression changes.

To further evaluate the visual quality of the generated face images, we design a user-study experiment that volunteers are asked to classify a given image as real or synthetic. We select 20 volunteers to perform the task, each one is shown a random selection of 50 real and 50 synthetic face images, and is asked to label the images as real or synthetic. The confusion matrix of the results is shown in Table~\uppercase\expandafter{\romannumeral2}. The volunteers choose the correct label 1048 times out of 2000 trials with a 52.4 \% probability, meaning the generated faces can be hardly differentiated from the real ones by human eyes.
\begin{table}
\small
\begin{center}
\caption{User-study experimental results. The average human classification accuracy is 52.4\%.}
\vspace{-0.55cm}
\resizebox{0.50 \textwidth}{!}{
\begin{tabular}{l l c c}
\hline
1 &   & Selected as real & Selected as synthetic \\
\hline
2 & Ground truth real & 513 & 487  \\
3 & Ground truth synthetic & 465 & 535  \\
\hline
\end{tabular}}
\end{center}
\label{tabxxx}
\end{table}
\subsection{Component Analysis}
We first investigate different settings and loss combinations of FaceAnime to see their respective roles in face generation. Specifically, we replace the sparse texture by the dense texture and facial landmarks to train our model. Two kinds of landmark priors are used: 1) 68 2D sparse facial landmarks and 2) 20,000+ 3D facial points. We also remove the identity-preserving loss $\mathcal{L}_{\text{\text{id}}}$ from the overall loss for training to evaluate how it can affect the generated images. We use a picture of Queen Elizabeth from the Internet to show the results of all the variants.

The visualization results are shown in the left panel of Figure~\ref{fig:7}. FaceAnime w/o $\mathcal{L}_{\text{\text{id}}}$ means removing $\mathcal{L}_{\text{\text{id}}}$ during training; FaceAnime-v1, -v2 and -v3 mean using the dense texture, 68 sparse 2D landmarks and dense 3D landmarks as a prior to guide face generation, respectively. Based on the observations, FaceAnime-v2 where the sparse 2D landmarks are used as a prior achieves the worst visual results. The generated image is blurred with most of the high-frequency details lost. When the dense 3D points are used as a prior (FaceAnime-v3), the visual quality of the result is much better than that of FaceAnime-v2. This is because the dense 3D points contain much more information such as facial shape, structural details and image depth than the sparse 2D landmarks, which leads to more realistic face generation. FaceAnime and FaceAnime-v1 (dense texture) achieve comparable visual results. By comparing them with FaceAnime-v2 and FaceAnime-v3, it is easy to see that involving the content information from the original image can greatly improve the visual quality for generated results. When the identity-preserving loss $\mathcal{L}_{\text{\text{id}}}$ is removed (see \textit{Col.} 2 of the left panel), we find the training is difficult to converge. The generated face contains visible structural information from the original face image, which cannot vanish even for a very long period of training. In addition, the identity of the generated face seems changed.

\begin{table}
\small
\begin{center}
\caption{Quantitative comparison results for FaceAnime variants on IJB-C testing set. FaceAnime, FaceAnime ($\bm{H}^{2}_{3d}$) and FaceAnime ($\bm{H}^{3}_{3d}$) represent the sparse texture mapping with an interval of 1, 2, and 3, respectively.}
\vspace{-0.2cm}
\resizebox{0.48 \textwidth}{!}{
\begin{tabular}{c l c c c c c}
\hline
1 & Variants \quad & PSNR $\uparrow$ & SSIM $\uparrow$ & FID $\downarrow$ & LRMS $\downarrow$   \\
\hline
2 & FaceAnime  & \textbf{29.87} & \textbf{0.65} & \textbf{49.10} & \textbf{0.21}\\
3 & FaceAnime ($\bm{H}^{2}_{3d}$) & 29.86 & 0.64 & 52.14 & 0.22\\
4 & FaceAnime ($\bm{H}^{3}_{3d}$) & 29.85 & 0.62 & 54.32 & 0.23\\
5 & FaceAnime  w/o $\mathcal{L}_{\text{\text{id}}}$ & 29.84 & 0.62 & 60.16 & 0.23\\
6 & FaceAnime-v1 & \textbf{29.87} & 0.64 & 51.25 & 0.22\\
7 & FaceAnime-v2 & 29.83 & 0.61 & 62.37 & 0.24\\
8 & FaceAnime-v3 & 29.84 & 0.63 & 58.43 & 0.23\\
\hline
\end{tabular}}
\end{center}
\label{tab2}
\end{table}

\begin{table}
\small
\begin{center}
\caption{The time cost for our FaceAnimes compared with Pix2PixHD, X2Face and FSGAN. `$\#$ Lms' indicates the number of facial landmarks.  `-' means that the texture mapping is not used for the corresponding method. The speed is tested on 2080ti GPU, E5 2687w CPU based system.}
\vspace{-0.2cm}
\resizebox{0.48 \textwidth}{!}{
\begin{tabular}{c l c c c c c}
\hline
1 & Model \quad & $\#$ Lms & Texture mapping time & Model running time    \\
\hline
2 & Pix2PixHD & 68 & - & 30.0 ms \\
3 & X2Face & 68 & - & 109.5 ms\\
4 & FSGAN & 68 & - & 30.4 ms \\
\hline
5 & FaceAnime & 14692 & 450 ms & 15.3 ms \\
6 & FaceAnime-v1 & 22038 & 1550 ms & 15.3 ms \\
7 & FaceAnime-v2 & 68 & - & 15.3 ms \\
8 & FaceAnime-v3 & 22038 & 1550 ms & 15.3 ms \\
\hline
\end{tabular}}
\end{center}
\label{tab3}
\end{table}

To evaluate the effectiveness of the proposed sparse texture mapping, we consider a special case where the eyes of the source face are closed. The comparison results are shown in the right panel of Figure~\ref{fig:7}. As can be seen, when using the dense texture as a prior, the eyes of the generated face look very strange and asymmetric, and involve many artifacts. However, the result of using sparse texture as a prior does not suffer from such problems. This is because the dense texture prior is too strong, making the model hardly adapt to small changes of the target motion. The quantitative results of the variants are shown in Table~\uppercase\expandafter{\romannumeral3} (following the same experimental setting as in Sec. 3.0.2.), where we can see FaceAnime achieves the best result among all the variants, further confirming the effectiveness of the components in our framework. FaceAnime (interval equals 1) performs slightly better than FaceAnime-v1, this is because sparse texture and dense texture achieve comparable overall visualization results, however for some of the special cases (such as the case when the eyes of the source face are closed), sparse texture performs better than dense texture on eye regions. The results of  FaceAnime ($\bm{H}^{2}_{3d}$) and FaceAnime ($\bm{H}^{3}_{3d}$) are worse than those of dense texture, the reason should be that too few texture features are included to guide image generation when interval equals 2 or 3, which can not guarantee a better visualization generation results.

In addition, computational efficiency is also a critical aspect in the comprehensive evaluation of image generation. To evaluate the computational efficiency, we report the model running time of our FaceAnimes and compare with other methods. The texture mapping time is also reported for different FaceAnime variants. The results are shown in Table~\uppercase\expandafter{\romannumeral4}. As we can see, the model runing time of our FaceAnimes are the lowest compared with other methods, about 2$\times$ faster than Pix2PixHD and FSGAN, and 7.3$\times$ faster than X2Face, respectively. This is because our CNN architecture is more advanced than the other methods. The major time consumption of our methods is texture mapping. However, it is clear to see our sparse texture mapping is much more efficient than the dense texture mapping. FaceAnime is 3.4$\times$ faster than FaceAnime-v1 and FaceAnime-v3 where dense texture mapping is used. Please note that the computation of texture mapping increases exponentially with the number of facial landmarks. This is the reason why the number of landmarks of FaceAnime-v1 is about twice of the number of FaceAnime, while it is 3.4$\times$ slower than FaceAnime.


\subsection{Discussion}
In this section, we discuss two experimental designs for our method, \textit{i.e.,} using an interpolation operation instead of LSTM module for video prediction, and employing rendering during the process of face generation.

\begin{figure}[!t]
\centering{\includegraphics[width = 8.5cm, height=7.2cm]{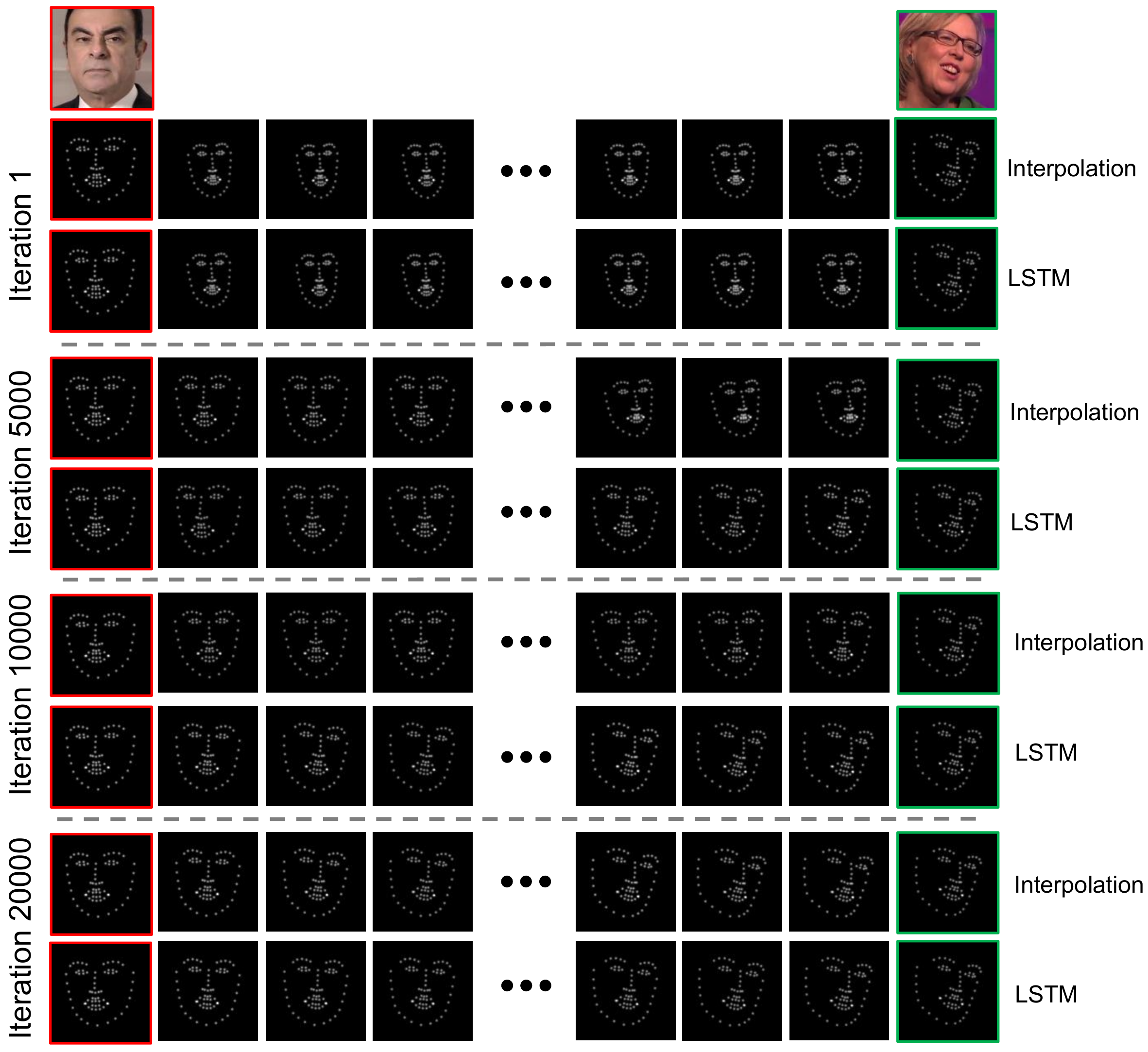}}
\vspace{-0.3cm}
\caption{The predicted facial landmarks by employing LSTM module and interpolation operation, respectively.
\label{fig:minor2}}
\end{figure}

For the target-driven prediction task, as the initial and target 3DMM parameters have been given, it is possible to use an interpolation operation instead of the LSTM module to train the prediction model. However, as can be seen from Figure~\ref{fig:minor2}, LSTM can converge well (the adjacent frames are consistent and smooth) at an iteration of 10, 000, while interpolation operation takes longer time to converge. Besides, LSTM is able to achieve consistent shape sequences at iteration 5, 000, while interpolation operation can not. The results reveal two strengths of LSTM over interpolation operation, \textit{i.e.}, faster convergence speed and better shape-preserved ability. Because LSTM is able to learn temporal correlations across frames during a long time interval, while interpolation would drop such temporal infomation. This continuous temporal information learned by LSTM could accelerate model convergence to a better state. In addition, the predicted 3DMM parameters can be decoded to restore 3D face shape, learning temporal correlations makes the restored face shapes more consistent along with time series.

In our work, texture is only used as the input of the network, serving as prior knowledge to guide face generation. Actually, feeding the rendering information as input is similar with image inpainting, where the residual region is taken as input to restore the missing regions. A number of face image inpainting works \cite{yeh2017semantic,zeng2019learning} have achieved good results only conditioned on the residual input. During our experiments, we find rendering for face generation would not help much to improve the image quality. In addition, 3D rendering is a pixel-by-pixel remapping process, which is time-consuming. Using the rendering operation during generation process would reduce the efficiency of the model. Please note that even the network is designed for a fixed size 128 $\times$ 128, it can also be designed for other image size, such as 256 $\times$ 256 and 512 $\times$ 512, just by increasing the size of convolution layer and sampling layer and deeper the network.

\section{Conclusion}
In this paper, we propose a novel framework for face video generation. The proposed method is a two-stage framework, where a source face image is firstly used to generate temporal 3D dynamics, which are rendered based on a proposed sparse texture mapping algorithm to carry face structural details and image content. Then the rendered sparse texture serves as strong priors for face generation. Different from most of previous methods that merely aim at face video generation, we focus on both face video generation and face video prediction. Three challenging tasks have been performed to show the effectiveness of our method, \textit{i.e.,} face video retargeting, video prediction and target-driven video prediction. The extensive experimental results have shown that our method is capable of retargeting the pose and expressions of a reference video to the source face, as well as predicting a stochastic and reasonable future for the source face or controlling the video generation towards a given target face. The qualitative and quantitative results have justified the superiority of our method in generating high-quality, visually pleasant, and identity-preserving faces over the other state-of-the-art face generation methods.

\section*{Acknowledgment}
This work was partially supported by the Open Fund Project of Key Laboratory of Flight Technology and Flight Safety of CAFUC (FZ2020KF10), the National Science Foundation of China (62006244), the Project of Comprehensive Reform of Electronic Information Engineering Specialty for Civil Aircraft Maintenance (14002600100017J172), the Project of Civil Aviation Flight University of China (Grant Nos. J2018-56,  CJ2019-03, J2020-060), and the Sichuan University Failure Mechanics \& Engineering Disaster Prevention and Mitigation Key Laboratory of Sichuan Province Open Foundation (2020FMSCU02).

\ifCLASSOPTIONcaptionsoff
  \newpage
\fi

\vspace{-1.2cm}
\begin{IEEEbiography}[{\includegraphics[width=1.1in,height=1.35in,clip,keepaspectratio]{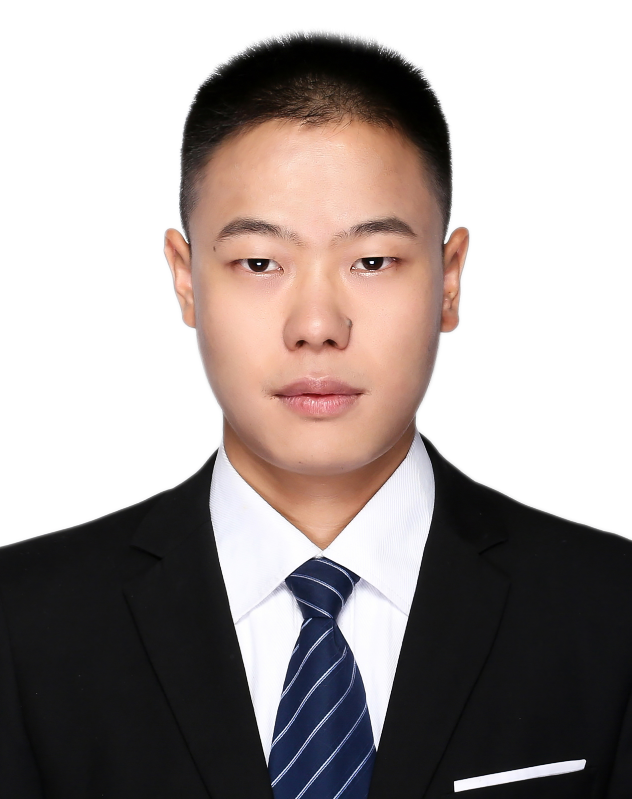}}]%
{Xiaoguang Tu}
is currently a lecturer in Aviation Engineering Institute at Civil Aviation Flight University of China. He received his Ph.D degree from the University of Electronic Science and Technology of China (UESTC) in 2020. He was a visiting scholar at Learning and Vision Lab, National University of Singapore (NUS) from 2018 to 2020 under the supervision of Dr. Jiashi Feng. His research interests include convex optimization, computer vision and deep learning.
\end{IEEEbiography}

\vspace{-1.2cm}

\begin{IEEEbiography}[{\includegraphics[width=1.1in,height=1.35in,clip,keepaspectratio]{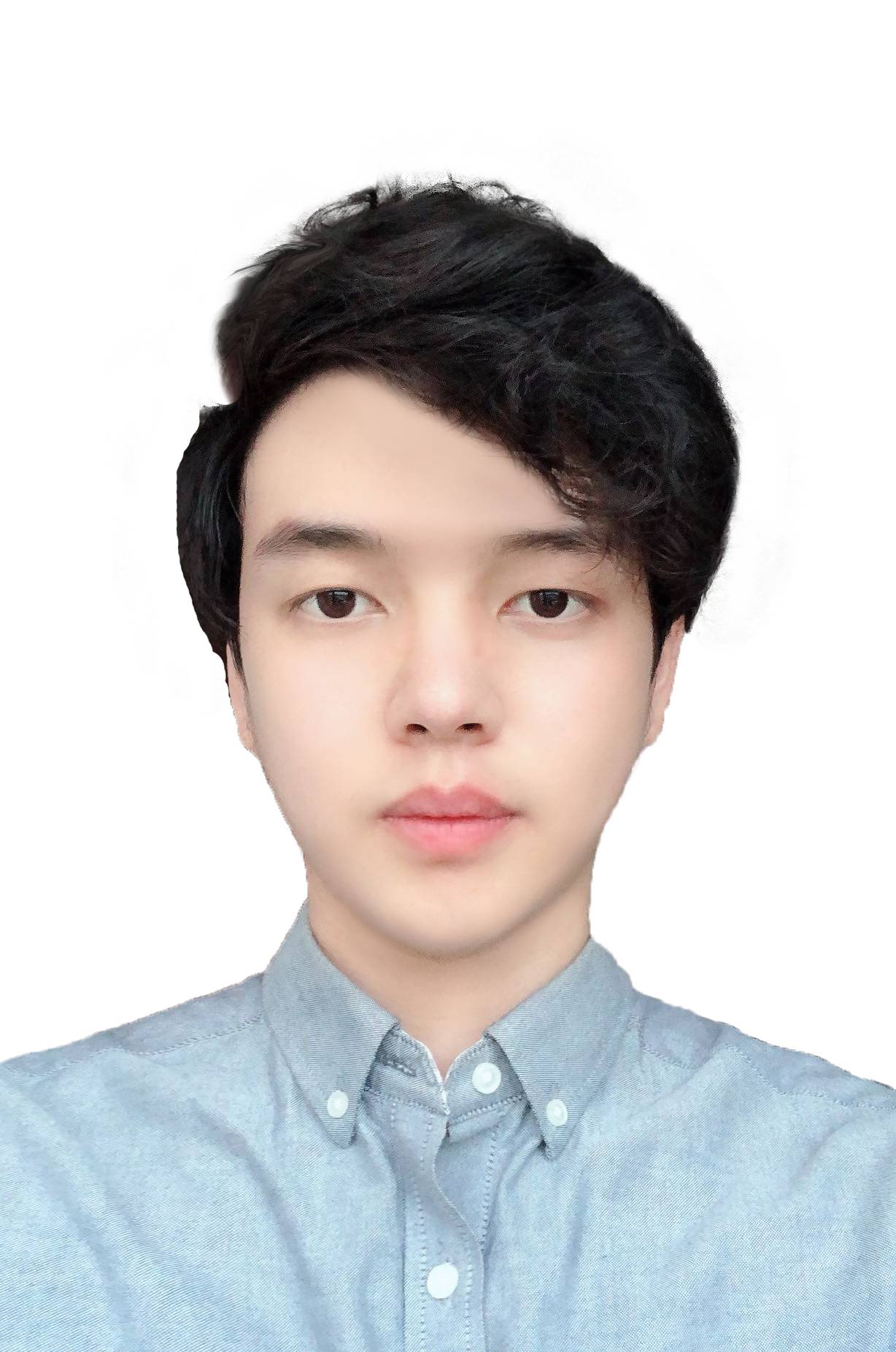}}]%
{Yingtian Zou}
is currently pursing the Ph.D in school of computing at National University of Singapore, Singapore. In 2018, He received his B.S in computer science from Huazhong University of Science and Technology, Wuhan, China. His research interests are computer vision, practical machine learning algorithms and its theory.
\end{IEEEbiography}

\vspace{-1.2cm}

\begin{IEEEbiography}[{\includegraphics[width=1.1in,height=1.35in,clip,keepaspectratio]{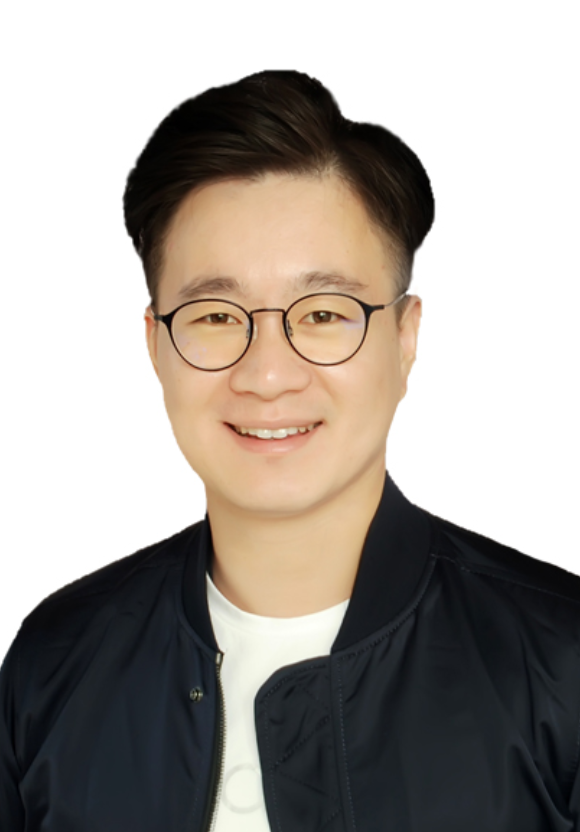}}]%
{Jian Zhao}
received the Bachelor degree from Beihang University in 2012, the Master degree from the National University of Defense Technology in 2014, and the Ph.D. degree from the National University of Singapore in 2019. He is currently an Assistant Professor with the Institute of North Electronic Equipment, Beijing, China. His main research interests include deep learning, pattern recognition, computer vision, and multimedia analysis. He has published over 40 cutting-edge papers. He has received the Young Talent Support Project from China Association for Science and Technology, and Beijing Young Talent Support Project from Beijing Association for Science and Technology, the Lee Hwee Kuan Award (Gold Award) on PREMIA 2019, the Best Student Paper Award on ACM MM 2018, and the top-3 awards several times on worldwide competitions. He is the SAC of VALSE, and the committee member of CSIG-BVD. He has served as the invited reviewer of NSFC, T-PAMI, IJCV, NeurIPS (one of the top 30\% highest-scoring reviewers of NeurIPS 2018), CVPR, etc.
\end{IEEEbiography}

\vspace{-1.2cm}

\begin{IEEEbiography}[{\includegraphics[width=1.1in,height=1.35in,clip,keepaspectratio]{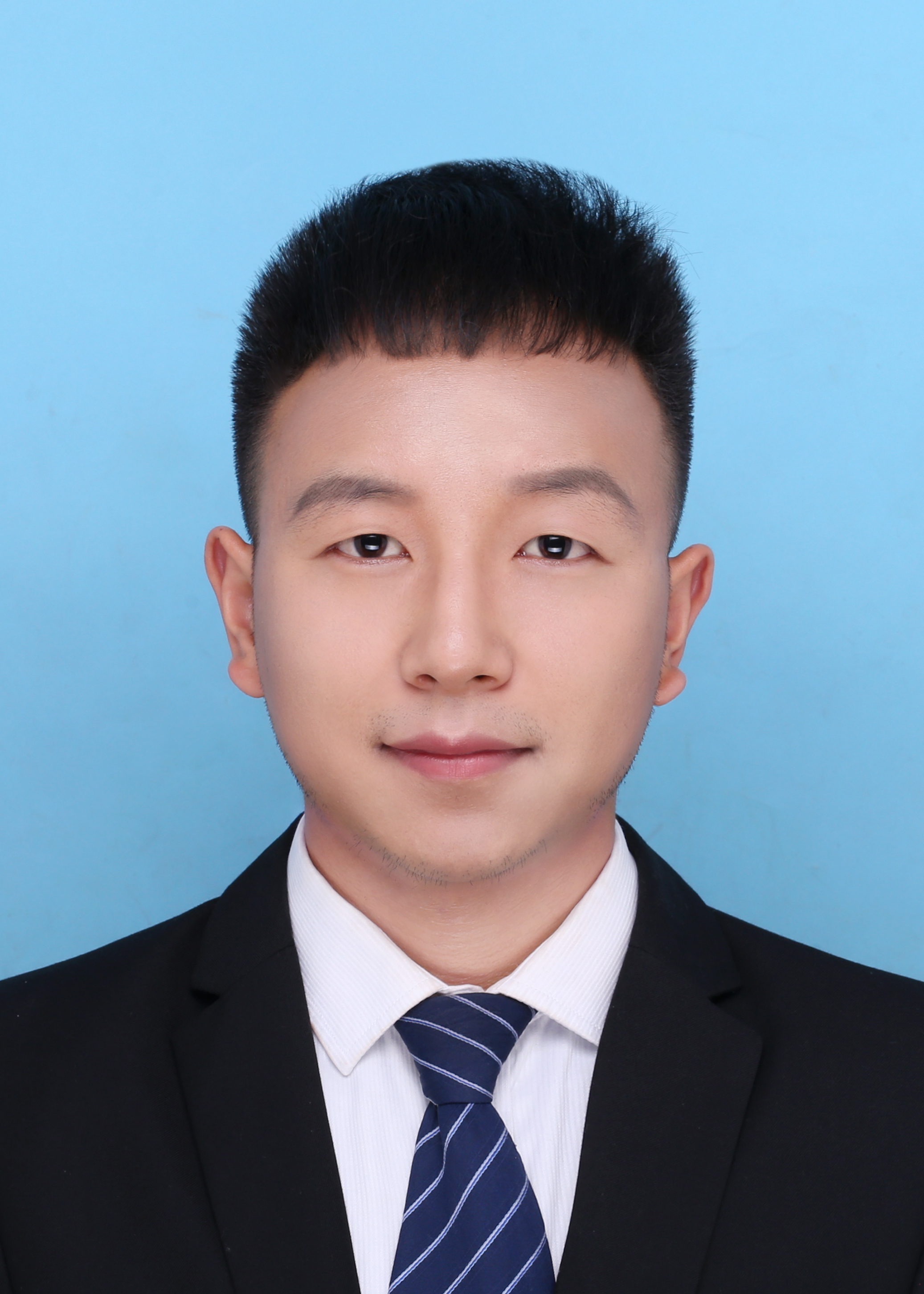}}]%
{Wenjie Ai}
is a Master with the School of Information and Communication Engineering at University of Electronic Science and Technology of China (UESTC). His research areas of interest mainly include computer vision and deep learning, in particular, super resolution and deblur.
\end{IEEEbiography}

\vspace{-1.2cm}

\begin{IEEEbiography}[{\includegraphics[width=1.1in,height=1.35in,clip,keepaspectratio]{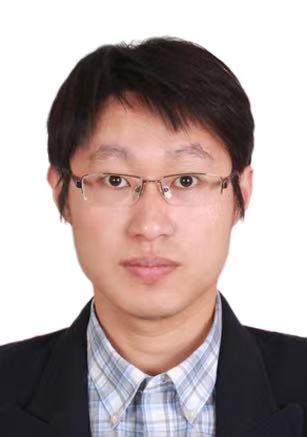}}]%
{Jian Dong}
(Member, IEEE) received the Ph.D. degree from the National University of Singapore. He is currently a director with Shopee. Prior to that, he was a senior director with Qihoo 360 and a research scientist with Amazon. His research interests include machine learning and computer vision. He received winner prizes in both PASCAL VOC and ILSVRC competitions..
\end{IEEEbiography}

\vspace{-0.7cm}

\begin{IEEEbiography}[{\includegraphics[width=1.1in,height=1.35in,clip,keepaspectratio]{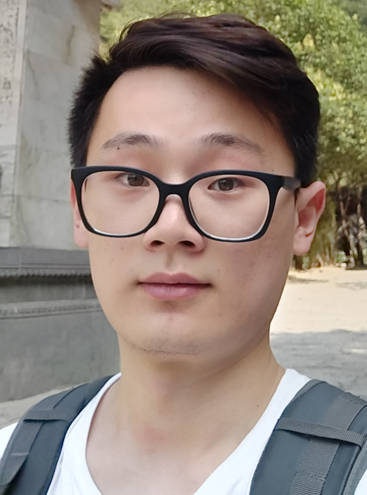}}]%
{Yuan Yao}
is an AI scientist at Pensees Pte. Ltd. He obtained his Master's degree in electrical and computer engineering at National University of Singapore in 2019. He was a visiting scholar at Cambridge Image Analysis Group, University of Cambridge in 2019 under the supervision of Prof. Carola-Bibiane Schonlieb. His research interests include generative adversarial networks, optical flow estimation and face recognition.
\end{IEEEbiography}

\vspace{-0.7cm}

\begin{IEEEbiography}[{\includegraphics[width=1.1in,height=1.35in,clip,keepaspectratio]{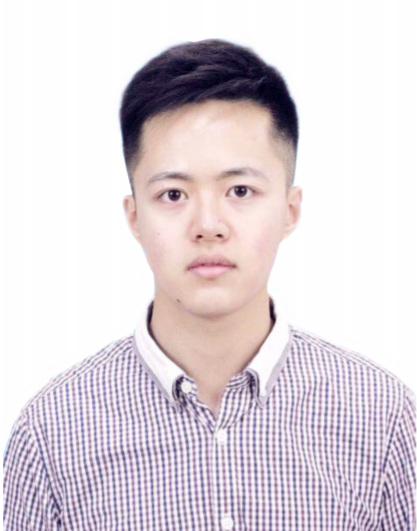}}]%
{Zhikang Wang}
is currently a master with the School of Electronic Engineering at Xidian University. He was a visiting scholar at Learning and Vision Lab, National University of Singapore (NUS) from 2019 to 2020 under the supervision of Dr. Jiashi Feng. He joined the Pensees Singapore as an intern in 2019. His research interests include computer vision, deep learning, and multimedia data processing.
\end{IEEEbiography}

\vspace{-0.7cm}

\begin{IEEEbiography}[{\includegraphics[width=1.1in,height=1.35in,clip,keepaspectratio]{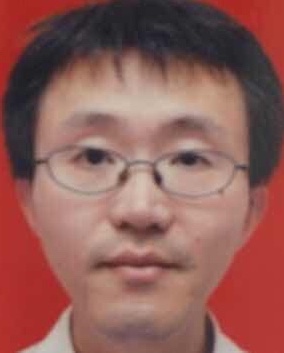}}]%
{Zhifeng Li}
is currently a top-tier principal researcher with Tencent AI Lab. He received the Ph. D. degree from the Chinese University of Hong Kong in 2006. After that, He was a postdoctoral fellow at the Chinese University of Hong Kong and Michigan State University for several years. Before joining Tencent AI Lab, he was a full professor with the Shenzhen Institutes of Advanced Technology, Chinese Academy of Sciences. His research interests include deep learning, computer vision and pattern recognition, and face detection and recognition. He is currently serving on the Editorial Boards of Neurocomputing and IEEE Transactions on Circuits and Systems for Video Technology. He is a fellow of British Computer Society (FBCS).
\end{IEEEbiography}

\vspace{-0.5cm}

\begin{IEEEbiography}[{\includegraphics[width=1.1in,height=1.35in,clip,keepaspectratio]{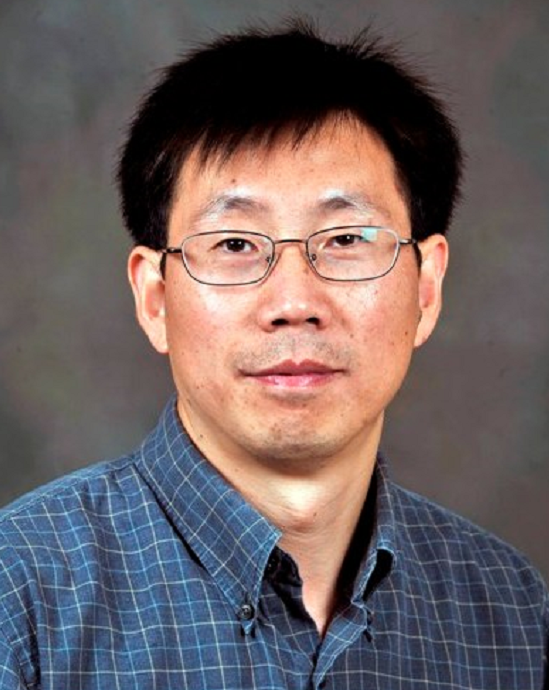}}]%
{Guodong Guo}
received the Ph.D. degree in computer science from University of Wisconsin, Madison, WI, USA. He is currently the Deputy Head of the Institute of Deep Learning, Baidu Research, and also an Associate Professor with the Department of Computer Science and Electrical Engineering, West Virginia University (WVU), USA. His research interests include computer vision, biometrics, machine learning, and multimedia. He received the North Carolina State Award for Excellence in Innovation in 2008, Outstanding Researcher (2017-2018, 2013-2014) at CEMR, WVU, and New Researcher of the Year (2010-2011) at CEMR, WVU. He was selected the ``People's Hero of the Week'' by BSJB under Minority Media and Telecommunications Council (MMTC) in 2013. Two of his papers were selected as ``The Best of FG'13" and ``The Best of FG'15", respectively.
\end{IEEEbiography}

\vspace{-0.5cm}

\begin{IEEEbiography}[{\includegraphics[width=1.1in,height=1.35in,clip,keepaspectratio]{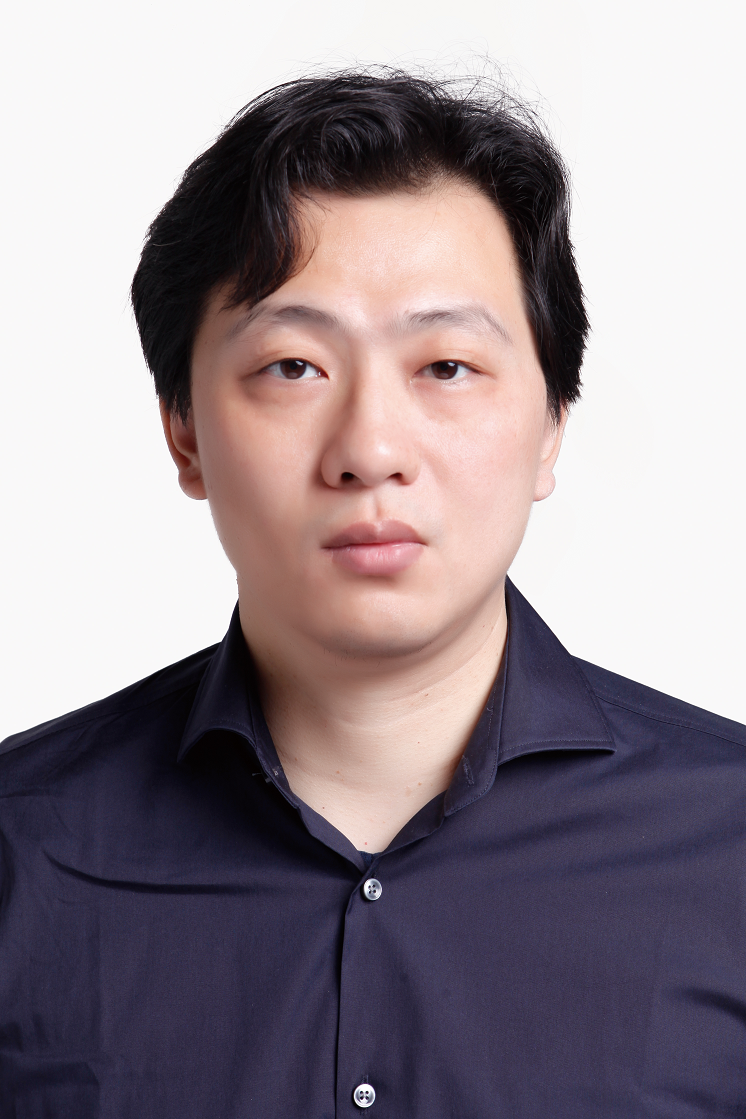}}]%
{Wei Liu}
is currently a Distinguished Scientist of Tencent, China and a director of Computer Vision Center at Tencent AI Lab.  Prior to that, he has been a research staff member of IBM T. J. Watson Research Center, Yorktown Heights, NY, USA from 2012 to 2015. Dr. Liu has long been devoted to research and development in the fields of machine learning, computer vision, pattern recognition, information retrieval, big data, etc. Dr. Liu currently serves on the editorial boards of IEEE Transactions on Pattern Analysis and Machine Intelligence, IEEE Transactions on Neural Networks and Learning Systems, IEEE Transactions on Circuits and Systems for Video Technology, Pattern Recognition, etc. He is a Fellow of the International Association for Pattern Recognition (IAPR) and an Elected Member of the International Statistical Institute (ISI).
\end{IEEEbiography}

\vspace{-0.5cm}

\begin{IEEEbiography}[{\includegraphics[width=1.1in,height=1.35in,clip,keepaspectratio]{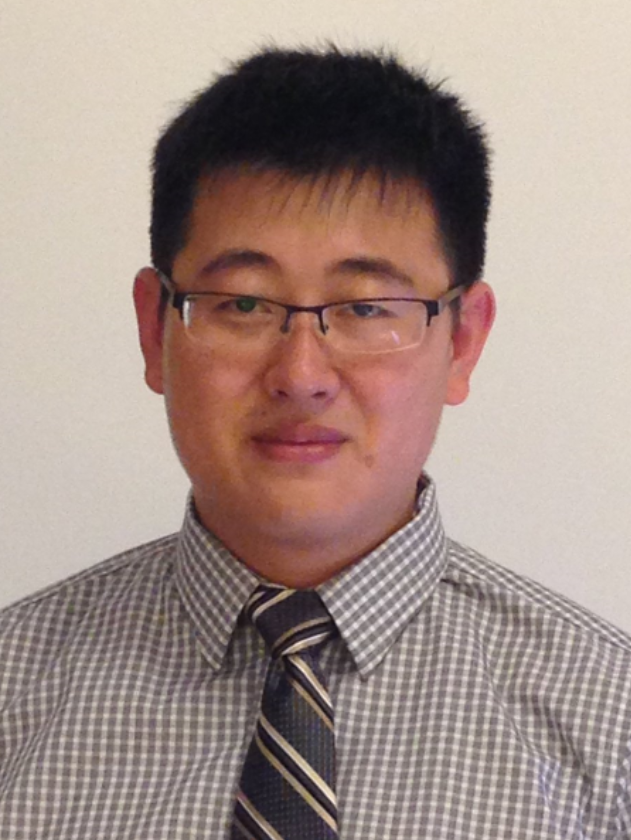}}]%
{Jiashi Feng}
received the B.E. degree from University of Science and Technology of China, Hefei, China, in 2007, and the Ph.D. degree from the National University of Singapore, Singapore, in 2014. He was a Post-Doctoral Researcher with the University of California from 2014 to 2015. He is currently an Assistant Professor with the Department of Electrical and Computer Engineering, National University of Singapore. His current research interests focus on machine learning and computer vision techniques for large-scale data analysis.
\end{IEEEbiography}

\end{document}